\documentclass[11pt,a4paper]{article}
\usepackage[T1]{fontenc}
\usepackage[utf8]{inputenc}
\usepackage{lmodern}
\usepackage[margin=25mm]{geometry}
\usepackage{amsmath,amssymb,amsthm}
\usepackage{microtype}
\usepackage{xcolor,comment}
\usepackage{tikz}
\usepackage[square,authoryear]{natbib}
\setcitestyle{aysep={,}}
\usepackage{fancyhdr}
\usepackage{hyperref}
\hypersetup{hypertexnames=false,colorlinks=true,linkcolor=black,citecolor=black,urlcolor=black,
  pdftitle={A Geometric Theory for Nonequilibrium Transformer Dynamics -- version 18},
  pdfauthor={Aythami Bethencourt de Leon; Darryl D. Holm},
  pdfkeywords={Transformer self-attention, Fisher-Rao gradient flows,
    relative entropy, tracking estimates, Poisson geometry, Hamiltonian dynamics}}
\allowdisplaybreaks[1]
\newcommand{\R}{\mathbb{R}}
\newcommand{\E}{\mathbb E}
\newcommand{\FR}{\mathrm{FR}}
\newcommand{\cF}{\mathcal{F}}
\newcommand{\tF}{\widetilde{\mathcal{F}}}
\DeclareMathOperator{\grad}{grad}
\DeclareMathOperator{\softmax}{softmax}
\DeclareMathOperator{\KL}{D_{KL}}
\newtheoremstyle{upright}
  {\topsep}{\topsep}{\normalfont}{}{\bfseries}{.}{.5em}{}
\theoremstyle{upright}
\newtheorem{theorem}{Theorem}[section]
\newtheorem{lemma}[theorem]{Lemma}
\newtheorem{proposition}[theorem]{Proposition}
\newtheorem{corollary}[theorem]{Corollary}
\theoremstyle{upright}
\newtheorem{definition}[theorem]{Definition}
\newtheorem{example}[theorem]{Example}
\theoremstyle{upright}
\newtheorem{remark}[theorem]{Remark}

\numberwithin{equation}{section}
\title{A Geometric Theory for Nonequilibrium Transformer Dynamics}
\author{Aythami Bethencourt de Le\'{o}n\thanks{Departamento de An\'{a}lisis Matem\'{a}tico, Universidad de La Laguna\quad
\href{mailto:abethenl@ull.edu.es}{abethenl@ull.edu.es}}\and Darryl D. Holm\thanks{Department of Mathematics, Imperial College London\quad
\href{mailto:d.holm@ic.ac.uk}{d.holm@ic.ac.uk}}}
\date{}
\begin{document}
\maketitle
\begin{center}
\end{center}
\begin{abstract}
We develop a variational and geometric theory based on statistical physics for
the
joint evolution of Transformer attention and token representations at fixed
model
parameters. To this end, we introduce a free energy for the attention and model
the evolution of the attention weights by its Fisher--Rao gradient flow.
At fixed token representations, we prove that the softmax matrix is the
unique free-energy minimiser and that every positive attention
trajectory converges to it. For prescribed time-dependent representations, we
show that the
attention dynamics becomes linear after a change to logit variables. This result
yields tracking bounds relative to the instantaneous softmax and sufficient
conditions for attention to converge to a fixed matrix.
Hamiltonian dynamics for the representations and their
momenta are then coupled to the attention relaxation
driven by the same energy. This coupling results in an energy-dissipation law.
We also develop two alternative Hamiltonians incorporating the attention value
map.
We prove global well-posedness of the attention flow and all three coupled
systems.
\end{abstract}
\tableofcontents

\section{Introduction}
Transformers underpin large language models (LLMs) and occupy a central place
in modern artificial intelligence \citep{Geshkovski2025}. Their attention
mechanism determines how token representations interact. Namely, pairwise scores
are converted by softmax into normalised weights, which are then used
to combine information from different positions \citep{Vaswani2017}.
As the token representations change, so do the scores and the resulting
interactions.

Here, we develop a variational and geometric theory
based on statistical physics for the joint
evolution of attention weights and token representations. We first model
attention as a Fisher--Rao gradient flow whose equilibrium solution
at fixed representations recovers softmax.
When the representations vary, attention retains
information from past scores and can depart from the current softmax weights.
To close the system, the attention dynamics is coupled to Hamiltonian
dynamics for the representations through a common energy.

\subsection{Framework and main results}
We represent each row of the attention matrix $A$ as a
probability distribution over source positions. We define a free energy
combining
the negative expected attention score with a negative-entropy term,
which penalises concentrated attention. For each
fixed representation matrix $X$, we prove that its unique minimiser is the
usual softmax matrix. We express the free-energy gap in terms of relative
entropy, whose local quadratic expansion motivates our choice of
the Fisher--Rao metric for the attention gradient flow.

For a prescribed $C^1$ representation path $X(t)$, we evolve
$A(t)$ by the Fisher--Rao gradient flow of the attention free energy.
We show that a change to logit variables linearises the dynamics and
results in an explicit solution. We prove global existence and uniqueness for
positive initial attention, whose positivity and row normalisation are
preserved. The solution
shows how the attention at a given time depends on an exponentially weighted
history
of scores. We bound attention's departure from the instantaneous
softmax and give sufficient conditions for convergence to a fixed attention
matrix.
In particular, for fixed $X$, every positive attention trajectory converges to
the softmax equilibrium.

To determine the joint dynamics of the representations $X(t)$ together with
attention,
we introduce canonical momenta $P(t)$ and a Hamiltonian comprising the sum of
kinetic
energy and the free energy of attention. Thus, we couple canonical Hamiltonian
dynamics
for $(X,P)$ to Fisher--Rao gradient descent of the same Hamiltonian in $A$.
To incorporate the attention value map into the representation
dynamics, we develop two alternative Hamiltonian formulations: one penalises
discrepancies between representations and attended values, while the other
sets $\dot X$ equal to the attention output. For all three systems, we derive an
energy-dissipation law and prove global existence and uniqueness for positive
initial attention, while preserving positivity and row normalisation.

\subsection{Relation to existing work}
Continuous-time models of self-attention treat token
representations as interacting particles, with attention weights computed
from their current features. This approach has revealed clustering and
metastability of the representations
\citep{Geshkovski2023,Geshkovski2024,Geshkovski2025}.
\citet{Burger2025} develop a Wasserstein-type gradient-flow
interpretation
of mean-field self-attention with normalised representations, under symmetry
and compatibility conditions on the interaction and value matrices.
Their geometry acts on distributions of representations on the sphere.
In our formulation, by contrast, the Fisher--Rao geometry acts on distributions
over
source positions, namely the attention rows. We give these rows their own
evolution and couple it to Hamiltonian dynamics for the representations
through a common energy. Our construction also requires no symmetry of the
query--key interaction matrix.

Our attention flow \eqref{eq:attention-flow1} is an
entropy-adjusted replicator equation in each row. \citet{Coucheney2015}
show that these equations are equivalent to accumulating payoffs with
exponential discounting and then applying a logit (softmax) choice map.
At the level of this row equation, source positions correspond to actions,
attention weights to action probabilities, and the scores $s_{ij}(X(t))$
to payoffs. \citet{MertikopoulosSandholm2018} place
replicator dynamics
in a broader class of game dynamics constructed from Riemannian metrics
on strategy simplices. The Shahshahani metric yielding replicator dynamics
in their framework is the Fisher--Rao metric used here.
In these game models, payoffs depend on strategy distributions. Our scores,
however, are generated by token representations $X(t)$, which we either
prescribe
or evolve together with attention and momenta through a common Hamiltonian.

\subsection{Scope and organisation}
{The structure of the paper is as follows.}

\begin{itemize}
\item
Section~\ref{sec:model} defines the attention model and free
energy, proves the softmax minimisation result, and motivates the
Fisher--Rao metric through relative entropy.
\item
Section~\ref{sec:preliminaries} introduces the geometric tools,
derives the Fisher--Rao gradient on the attention space, and relates
Fisher--Rao gradient flows to logit dynamics.
\item
Section~\ref{sec:attention-flow} derives the explicit attention
solution for prescribed representations, establishes global well-posedness
and tracking estimates, and studies convergence and the free-energy balance.
\item
Section~\ref{sec:hamiltonian} develops the canonical Hamiltonian
coupling and two alternatives incorporating the value map, derives their
energy-dissipation laws, and proves global well-posedness for all three systems.
\item
Section~\ref{sec:summary} summarises the results and discusses
directions for further work.
\end{itemize}

Appendix~\ref{sec:geometry-details} collects supporting geometric
proofs and entropy interpretations of the Fisher--Rao metric.
{Appendix~\ref{sec:depth} defines the Transformer residual blocks and
establishes the one-step consistency of their scaled updates with a
continuous-depth equation.}

\section{The attention model}\label{sec:model}\label{sec:free-energy}
\subsection{Features and scores}\label{sec:model-scores}
{A Transformer processes a sequence of $N\geq2$ tokens. For example,
a language model splits an input text into tokens, such as words, parts
of words or punctuation, and represents each token by a vector in a
feature space $\R^d$, with $d\geq1$. Successive Transformer layers update
these vectors using information from the token sequence. We call them
token representations or feature vectors.}

{At a given Transformer layer, we collect the current representations
of the token sequence in} $X=(x_1,\ldots,x_N)\in\mathcal X:=\R^{d\times N}$,
where column $i$ is the feature vector of token $i$, for
$i=1,\ldots,N$. {Fix a query/key dimension $d_k\geq1$.}
We consider a single unmasked attention head, with matrices
\[
W_Q,W_K\in\R^{d_k\times d},\qquad W_V\in\R^{d\times d}.
\]
These matrices remain fixed throughout the attention and
Hamiltonian models developed in the main text.
\begingroup
At that layer, the attention mechanism defines the query, key and value
at position $i$, respectively, by
\[
q_i=W_Qx_i\in\R^{d_k},\qquad k_i=W_Kx_i\in\R^{d_k},\qquad v_i=W_Vx_i\in\R^d,
\qquad i=1,\ldots,N.
\]
Informally, the query $q_i$ describes what information the position $i$
seeks, the key $k_i$ provides a descriptor against which queries are matched,
and the value $v_i$ carries the information to be combined in the output.
The scaled inner product of the query at position $i$ and the key at
position $j$ defines their attention score $s_{ij}(X)$.
\endgroup
{Collecting these scores gives the map}
$S:\mathcal X\to\R^{N\times N}$, defined by
\begin{equation*}
\begin{aligned}
M&:=\frac{W_Q^\top W_K}{\sqrt{d_k}},\qquad
S(X):=X^\top MX,\\
s_{ij}(X)&=x_i^\top Mx_j=\frac{q_i^\top k_j}{\sqrt{d_k}},
\qquad i,j=1,\ldots,N.
\end{aligned}
\end{equation*}
Here $i$ is called the receiving position and $j$ is called the source position.
{For a fixed temperature $\tau>0$, softmax converts each row of scores
into attention weights:}
\begin{equation}\label{eq:standard-attention}
A^*_{ij}(X)=\frac{\exp(s_{ij}(X)/\tau)}
{\sum_{k=1}^N\exp(s_{ik}(X)/\tau)},
\qquad i,j=1,\ldots,N.
\end{equation}
{For fixed scores, lower temperatures $\tau$ favour the largest scores more
strongly, while higher temperatures give more uniform weights.}

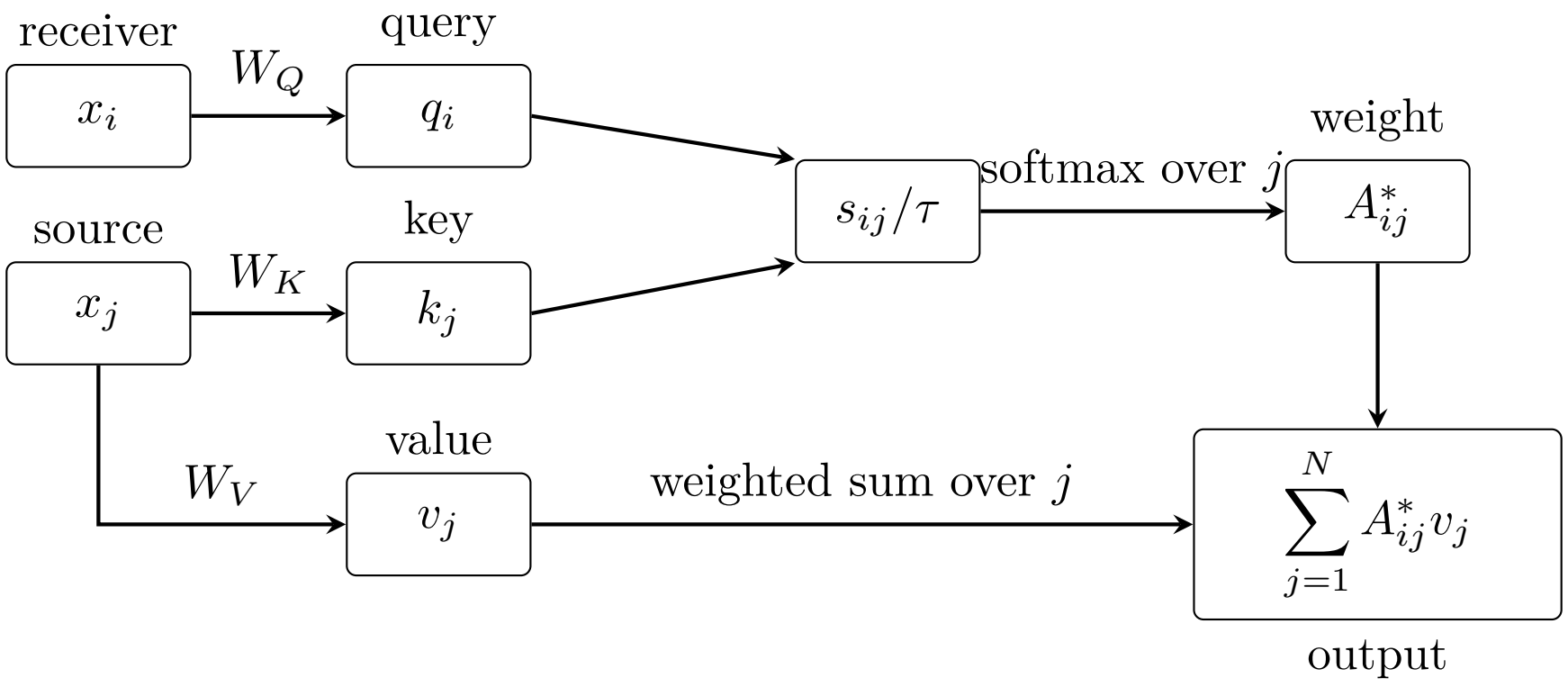
\begin{figure}[ht]
\centering
\begingroup
\colorlet{attentiondiagramcolor}{.}
\begin{tikzpicture}[x=1cm,y=1cm,>=stealth,
  every node/.style={font=\small},
  representation/.style={draw=attentiondiagramcolor,rounded corners=2pt,
    minimum width=1.35cm,minimum height=.75cm,align=center},
  flow/.style={->,draw=attentiondiagramcolor,thick}]
\node[representation] (xi) at (0,1.2) {$x_i$};
\node[above] at (xi.north) {receiver};
\node[representation] (xj) at (0,-.25) {$x_j$};
\node[above] at (xj.north) {source};
\node[representation] (qi) at (2.5,1.2) {$q_i$};
\node[above] at (qi.north) {query};
\node[representation] (kj) at (2.5,-.25) {$k_j$};
\node[above] at (kj.north) {key};
\node[representation] (vj) at (2.5,-1.8) {$v_j$};
\node[above] at (vj.north) {value};
\node[representation] (score) at (5.8,.5) {$s_{ij}/\tau$};
\node[representation] (weight) at (9.4,.5) {$A^*_{ij}$};
\node[above] at (weight.north) {weight};
\node[representation,minimum width=2.7cm] (output) at (9.4,-1.8)
  {$\displaystyle\sum_{j=1}^N A^*_{ij}v_j$};
\node[below] at (output.south) {output};
\draw[flow] (xi)--node[above] {$W_Q$} (qi);
\draw[flow] (xj)--node[above] {$W_K$} (kj);
\draw[flow] (xj.south)--(0,-1.8)--node[above] {$W_V$} (vj.west);
\draw[flow] (qi.east)--(score.north west);
\draw[flow] (kj.east)--(score.south west);
\draw[flow] (score)--node[above,align=center] {softmax over $j$} (weight);
\draw[flow] (weight)--(output);
\draw[flow] (vj)--node[above] {weighted sum over $j$} (output);
\end{tikzpicture}
\endgroup
\caption{{Self-attention for a fixed receiving position $i$.
Its query is compared with the keys at all source positions
$j=1,\ldots,N$. Softmax converts the scores into weights, which are
then used to combine the value vectors.}}
\label{fig:attention-construction}
\end{figure}

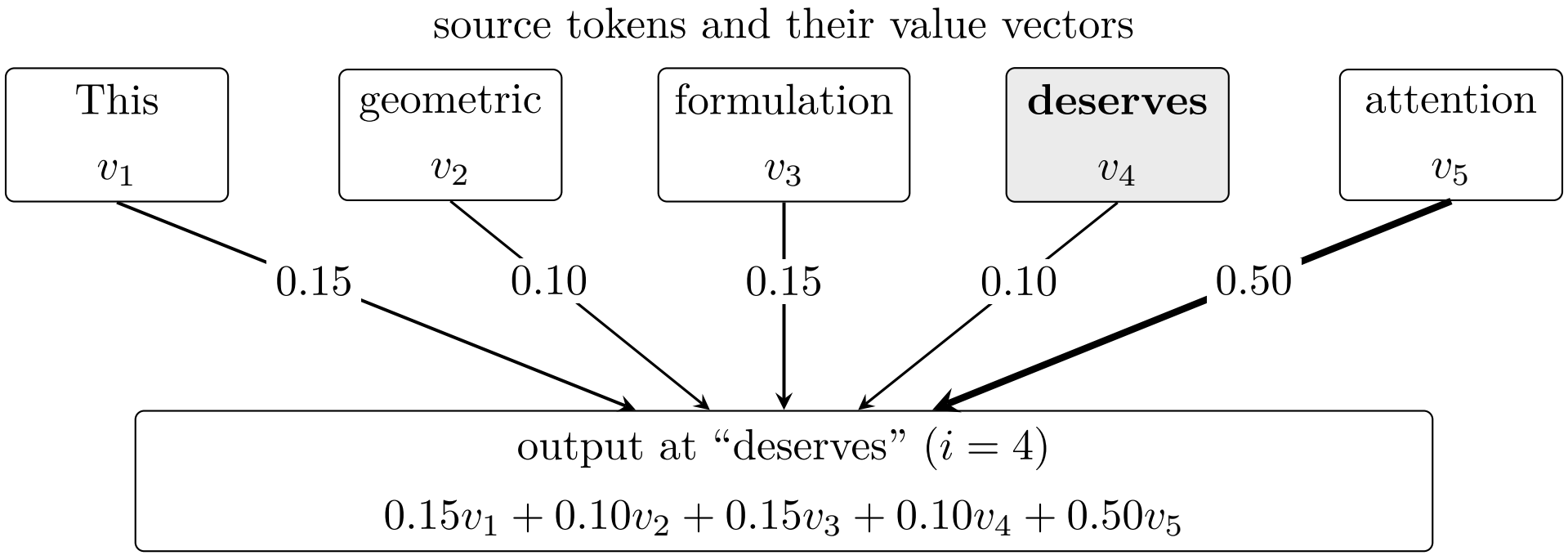
\begin{figure}[ht]
\centering
\begingroup
\colorlet{attentionexamplecolor}{.}
\begin{tikzpicture}[x=1cm,y=1cm,>=stealth,
  every node/.style={font=\small},
  source/.style={draw=attentionexamplecolor,rounded corners=2pt,
    minimum width=1.8cm,minimum height=.95cm,align=center},
  weight/.style={fill=white,inner sep=2pt},
  flow/.style={->,draw=attentionexamplecolor}]
\node[above] at (5.4,.65) {source tokens and their value vectors};
\node[source] (v1) at (0,0) {This\\[3pt]$v_1$};
\node[source] (v2) at (2.7,0)
  {geometric\\[3pt]$v_2$};
\node[source] (v3) at (5.4,0)
  {formulation\\[3pt]$v_3$};
\node[source,fill=attentionexamplecolor!8] (v4) at (8.1,0)
  {\textbf{deserves}\\[3pt]$v_4$};
\node[source] (v5) at (10.8,0)
  {attention\\[3pt]$v_5$};
\node[source,minimum width=10.5cm,minimum height=1.1cm] (output)
  at (5.4,-2.8) {output at ``deserves'' ($i=4$)\\[4pt]
  $0.15v_1+0.10v_2+0.15v_3+0.10v_4+0.50v_5$};
\draw[flow,line width=.75pt] (v1.south)--node[weight,pos=.38] {$0.15$}
  ([xshift=-1.2cm]output.north);
\draw[flow,line width=.6pt] (v2.south)--node[weight,pos=.38]
  {$0.10$} ([xshift=-.6cm]output.north);
\draw[flow,line width=.75pt] (v3.south)--node[weight,pos=.38]
  {$0.15$} (output.north);
\draw[flow,line width=.6pt] (v4.south)--node[weight,pos=.38]
  {$0.10$} ([xshift=.6cm]output.north);
\draw[flow,line width=1.6pt] (v5.south)--node[weight,pos=.38] {$0.50$}
  ([xshift=1.2cm]output.north);
\end{tikzpicture}
\endgroup
\caption{An illustrative attention row for the phrase
``This geometric formulation deserves attention'',
treating each word as one token. The receiving position is
$i=4$; the arrow labels give the chosen weights
$A^*_{4j}$.}
\label{fig:attention-example}
\end{figure}

{For each receiving position $i$, the weights $A^*_{ij}(X)$ sum to one over
source positions,
and the output is the weighted average of the values,
$\sum_{j=1}^N A^*_{ij}(X)v_j$. Since attention is unmasked, all positions,
including $i$ itself, contribute. Figure~\ref{fig:attention-construction}
summarises the attention computation, and Figure~\ref{fig:attention-example}
illustrates one receiving row. Appendix~\ref{sec:depth} describes the
residual blocks that update $X$ and their continuous-depth scaling.}

\subsection{Independent attention and its energy}\label{sec:independent-attention}
We allow attention to adjust over time instead of imposing
\eqref{eq:standard-attention} at each instant. We consider the following
state space.
\begin{definition}[Attention state]\label{def:attention-state}
\begin{equation*}
\mathcal A=\left\{A\in\R^{N\times N}: A_{ij}>0,\quad
\sum_{j=1}^NA_{ij}=1\ \text{for every }i\right\}.
\end{equation*}
\end{definition}
Each row of an attention matrix $A\in\mathcal A$ is a
strictly positive probability vector: its entries are positive and sum to one.
It therefore specifies a
probability distribution over source positions.
Formula~\eqref{eq:standard-attention} defines the map
$A^*:\mathcal X\to\mathcal A$.
For the independent state $A$, the feature attention output is $W_VXA^\top$.
The star in $A^*(X)$ distinguishes the algebraic softmax weights from
$A$, which we will model as a dynamical variable.

For each receiving position $i=1,\ldots,N$, define
\[
a^{(i)}=(A_{i1},\ldots,A_{iN})^\top,\qquad
s_i(X)=(s_{i1}(X),\ldots,s_{iN}(X))^\top.
\]
For a source-dependent quantity $b=(b_1,\ldots,b_N)$, define its
expectation under attention row $i$ by
\[
\E_i[b]:=\sum_{j=1}^N A_{ij}b_j.
\]
Logarithms of vectors are taken entrywise.

\begin{definition}[Attention free energy]\label{def:attention-energy}
For $A\in\mathcal A$ and $X\in\mathcal X$, define the attention free energy by
\begin{equation}\label{eq:total-free-energy}
\begin{aligned}
\cF(A;X)
&=-\sum_{i,j=1}^NA_{ij}s_{ij}(X)
+\tau\sum_{i,j=1}^NA_{ij}\log A_{ij}\\
&=\sum_{i=1}^N\left(-\E_i[s_i(X)]
+\tau\E_i[\log a^{(i)}]\right).
\end{aligned}
\end{equation}
\end{definition}
For each receiving row $i\in\{1,\ldots,N\}$, minimising the score term
$-\E_i[s_i(X)]$ is equivalent to maximising the attention-weighted score
$\E_i[s_i(X)]$. The term $\tau\E_i[\log a^{(i)}]$ is minus $\tau$ times
the row entropy, so minimising it favours higher entropy and discourages
concentrated attention. It acts as an entropic regulariser, with $\tau$
controlling its strength relative to score maximisation.

Consider now each term on its own, with $X$ fixed. The score term has
infimum $-\max_{1\leq j\leq N}s_{ij}(X)$, approached by concentrating
attention on sources with the largest score.
The entropy term alone is uniquely minimised by the uniform row
$A_{ij}=1/N$ for $j=1,\ldots,N$, with value $-\tau\log N$.

\subsection{Variational softmax}\label{sec:variational-softmax}
The equilibrium state of the free energy may be characterised by relative
entropy.
\begin{definition}[Relative entropy]\label{def:prelim-kl}
Let $n\geq2$ and let $p,q\in(0,\infty)^n$ satisfy
$\sum_{j=1}^n p_j=\sum_{j=1}^n q_j=1$. Define relative entropy $\KL(p\|q)$ as
\begin{equation*}
\KL(p\|q):=\sum_{j=1}^n p_j\log\frac{p_j}{q_j}.
\end{equation*}
\end{definition}
For outcomes drawn from $p$, relative entropy measures the average extra
information required by an ideal encoding based on $q$ rather than $p$.
With natural logarithms, it is measured in nats.
It is nonnegative, with equality only for $p=q$, but is generally
asymmetric and is not a metric; see Appendix~\ref{app:kl}
for the nonnegativity proof.

\begin{proposition}[Softmax minimises the total free energy]
\label{prop:softmax-minimiser}
Fix $X\in\mathcal X$ and $\tau>0$. The matrix
\[
A^*_{ij}(X)=\frac{e^{s_{ij}(X)/\tau}}{Z_i(X)},
\qquad i,j=1,\ldots,N,
\qquad\text{where } Z_i(X)=\sum_{k=1}^N e^{s_{ik}(X)/\tau},
\]
is the unique minimiser of $\cF(\cdot;X)$ on $\mathcal A$.
Writing $a^{(i)}=(A_{i1},\ldots,A_{iN})^\top$ and
$a^{*(i)}=(A^*_{i1},\ldots,A^*_{iN})^\top$ for the corresponding
rows as column vectors, one has
\begin{align}
\cF(A^*(X);X)&=-\tau\sum_{i=1}^N\log Z_i(X),
\nonumber\\
\cF(A;X)-\cF(A^*(X);X)
&=\tau\sum_{i=1}^N\KL\bigl(a^{(i)}\|a^{*(i)}\bigr).
\label{eq:free-energy-gap}
\end{align}
\end{proposition}
\begin{proof}
By \eqref{eq:total-free-energy}, the objective and its constraints
separate by rows. For each receiving row $i\in\{1,\ldots,N\}$, define
its contribution to the free energy on the positive orthant by
\[
\cF_i(a;X):=\sum_{j=1}^N a_j[-s_{ij}(X)+\tau\log a_j],
\qquad a\in(0,\infty)^N.
\]
Then $\cF(A;X)=\sum_{i=1}^N\cF_i(a^{(i)};X)$.
Fix $i$ for the row calculation and write $s_j=s_{ij}(X)$ and
$\pi_j=A^*_{ij}(X)$. To find the stationary points of $\cF_i$ under
the constraint $\sum_{j=1}^N a_j=1$, apply Lagrange multipliers to
\[
\tF_i(a,\lambda_i;X)=\cF_i(a;X)
+\lambda_i\left(\sum_{j=1}^N a_j-1\right).
\]
Its stationary equations are
\begin{equation*}
-s_j+\tau(1+\log a_j)+\lambda_i=0\quad(j=1,\ldots,N),
\qquad \sum_{j=1}^N a_j=1.
\end{equation*}
The first equations give
\[
\log a_j=\frac{s_j}{\tau}-1-\frac{\lambda_i}{\tau},
\qquad j=1,\ldots,N.
\]
Exponentiating yields $a_j=C_i e^{s_j/\tau}$, where
$C_i=e^{-1-\lambda_i/\tau}$ is independent of $j$.
With $Z_i=\sum_{j=1}^N e^{s_j/\tau}$, normalisation gives
\[
1=\sum_{j=1}^N a_j=C_i\sum_{j=1}^N e^{s_j/\tau}=C_iZ_i.
\]
Hence $C_i=Z_i^{-1}$ and $\lambda_i=\tau(\log Z_i-1)$.
The stationary point is therefore the softmax row $a=\pi$
from \eqref{eq:standard-attention}.

To establish global minimality, we substitute
$\log\pi_j=s_j/\tau-\log Z_i$ in the definition of $\cF_i$ above.
Since $\sum_{j=1}^N\pi_j=1$, we have
\[
\cF_i(\pi;X)=\sum_{j=1}^N\pi_j(-s_j+\tau\log\pi_j)
=-\tau\log Z_i\sum_{j=1}^N\pi_j=-\tau\log Z_i.
\]
For any positive probability vector $a$, the same identity and
Definition~\ref{def:prelim-kl} give
\begin{align*}
\cF_i(a;X)-\cF_i(\pi;X)
&=\tau\sum_{j=1}^N a_j
\left(\log a_j-\frac{s_j}{\tau}+\log Z_i\right)\\
&=\tau\sum_{j=1}^N a_j\log\frac{a_j}{\pi_j}
=\tau\KL(a\|\pi).
\end{align*}
Proposition~\ref{prop:prelim-kl} gives nonnegativity of this gap,
with equality only for $a=\pi$. Summing over receiving rows proves
both identities in \eqref{eq:free-energy-gap} and uniqueness of the
minimiser on $\mathcal A$.
\end{proof}

\subsection{From free energy to Fisher--Rao geometry}
\label{sec:energy-geometry}
For fixed $X\in\mathcal X$, \eqref{eq:free-energy-gap} expresses
$\cF(A;X)-\cF(A^*(X);X)$ as
$\tau\sum_{i=1}^N\KL(a^{(i)}\|a^{*(i)})$.
A variation $A_\varepsilon=A^*(X)+\varepsilon B$ preserves row
normalisation precisely when
\[
\sum_{j=1}^N B_{ij}=0,\qquad i=1,\ldots,N,
\]
since each row of $A^*(X)$ already sums to one.
For such a matrix $B\in\R^{N\times N}$, strict positivity of $A^*(X)$
ensures that $A_\varepsilon\in\mathcal A$ for sufficiently small
$|\varepsilon|$. Write $b^{(i)}=(B_{i1},\ldots,B_{iN})^\top$.
Using the local expansion from Proposition~\ref{prop:prelim-kl}
in Appendix~\ref{app:kl}, we obtain
\begin{equation}\label{eq:free-energy-quadratic}
\begin{aligned}
\cF(A_\varepsilon;X)-\cF(A^*(X);X)
&=\tau\sum_{i=1}^N
\KL\bigl(a^{*(i)}+\varepsilon b^{(i)}\|a^{*(i)}\bigr)\\
&=\frac{\tau\varepsilon^2}{2}
\sum_{i=1}^N g_{a^{*(i)}}(b^{(i)},b^{(i)})+O(\varepsilon^3)\\
&=\frac{\tau\varepsilon^2}{2}
\sum_{i,j=1}^N\frac{B_{ij}^2}{A^*_{ij}(X)}+O(\varepsilon^3)\\
&=\frac{\tau\varepsilon^2}{2}
g_{A^*(X)}(B,B)+O(\varepsilon^3),
\end{aligned}
\end{equation}
where $g_p(v,v)=\sum_{j=1}^N v_j^2/p_j$ for a positive probability
vector $p\in\R^N$ and $v\in\R^N$ with ${\sum_{j=1}^N v_j=0}$.
This is the Fisher--Rao squared length for one probability vector, and
\[
g_{A^*(X)}(B,B):=\sum_{i=1}^N
 g_{a^{*(i)}}(b^{(i)},b^{(i)})
 =\sum_{i,j=1}^N\frac{B_{ij}^2}{A^*_{ij}(X)}
\]
is the corresponding squared length on the attention space.
Section~\ref{sec:preliminaries} constructs these metrics; the full
attention metric is given in \eqref{eq:attention-metric}.
Thus the leading term of the free-energy gap is proportional to the
squared Fisher--Rao length of the variation. This motivates our choice
of the Fisher--Rao gradient flow for attention relaxation.

\section{Geometry of attention}\label{sec:preliminaries}
The expression $g_{A^*(X)}(B,B)$ in
\eqref{eq:free-energy-quadratic} motivates using the Fisher--Rao metric
to measure changes of attention. We now construct
this metric and its gradient on the attention space $\mathcal A$.
The geometric constructions below are classical;
Appendix~\ref{sec:geometry-details} collects their supporting proofs and
entropy interpretations.
The Fisher--Rao conventions follow \citet{AmariNagaoka2000} and
\citet{Amari2016}.

\subsection{Riemannian metrics and gradients}\label{sec:prelim-gradient}
A smooth state space is represented by a manifold $M$. Its tangent
space $T_pM$ consists of tangent vectors: the instantaneous
velocities of curves through $p\in M$. A Riemannian metric is a smoothly
varying inner product on each $T_pM$. It generalises the Euclidean dot product and makes lengths,
angles and gradients depend on the state $p\in M$.

\begin{definition}[Riemannian metric]\label{def:prelim-metric}
Let $M$ be a smooth finite-dimensional manifold. A Riemannian metric $g$
assigns to each $p\in M$ a symmetric bilinear form
$g_p:T_pM\times T_pM\to\R$ such that $g_p(v,v)>0$ for every nonzero
$v\in T_pM$, and the assignment is smooth in local coordinates.
\end{definition}
We write $\|v\|_{g,p}^2=g_p(v,v)$. In a coordinate chart
$z=(z^1,\ldots,z^r):U\to z(U)\subset\R^r$, with
$U\subset M$ and $r=\dim M$, the metric at $p\in U$ is represented by
\[
G(z(p))=(g_{\alpha\beta}(p))_{\alpha,\beta=1}^r,
\qquad
g_{\alpha\beta}(p)
=g_p\bigl(\partial_{z^\alpha}|_p,\partial_{z^\beta}|_p\bigr).
\]
Thus $G$ is symmetric and positive definite at every point, and its
entries depend smoothly on the coordinates.

\begin{definition}[Metric identifications]\label{def:musical}
Let $T_p^*M$ denote the dual vector space of $T_pM$.
For each $p\in M$, define
\[
g_p^\flat:T_pM\longrightarrow T_p^*M,\qquad
v\longmapsto g_p(v,\cdot),\qquad g_p^\sharp:=(g_p^\flat)^{-1}.
\]
Bilinearity makes $g_p^\flat$ linear, and positive
definiteness makes it injective. Therefore, since $T_pM$ and $T_p^*M$ have the same
finite dimension, it is an isomorphism. We also write
$v^\flat=g_p^\flat(v)$ and $\alpha^\sharp=g_p^\sharp(\alpha)$
for $v\in T_pM$ and $\alpha\in T_p^*M$.
\end{definition}
For a $C^1$ function $F:M\to\R$, the differential
$dF_p:T_pM\to\R$ is the linear map defined by
\[
dF_p(v)=\left.\frac{d}{ds}F(\sigma(s))\right|_{s=0},
\qquad \sigma(0)=p,\quad \dot\sigma(0)=v,
\]
where $\sigma:(-\varepsilon,\varepsilon)\to M$, $\varepsilon>0$, is any
$C^1$ curve with the indicated initial point and velocity.
On an open set $U\subset\R^r$, the Euclidean gradient of a $C^1$
function $F:U\to\R$ is characterised by the identity
\[
dF_x(v)=\nabla F(x)\cdot v,
\qquad x\in U,\quad v\in\R^r.
\]
The Riemannian gradient extends this identity by replacing the
Euclidean dot product with the metric.

\begin{definition}[Riemannian gradient]\label{def:prelim-gradient}
Let $(M,g)$ be a Riemannian manifold and $F:M\to\R$ a $C^1$ function.
Its Riemannian gradient is the vector field $\grad_g F$
characterised by
\begin{equation*}
g_p(\grad_g F(p),v)=dF_p(v)
\qquad\text{for every }p\in M\text{ and }v\in T_pM.
\end{equation*}
Equivalently, $\grad_g F(p)=g_p^\sharp(dF_p)=(dF_p)^\sharp$.
\end{definition}

In the local coordinates above, applying the defining
identity to each coordinate tangent vector gives
\begin{equation}
[\grad_g F(p)]_z=G(z(p))^{-1}\nabla\widehat F(z(p)),
\qquad \widehat F:=F\circ z^{-1}.
\label{eq:prelim-coordinate-gradient}
\end{equation}
In particular, $\grad_g F$ is continuous when $F$ is $C^1$.

\subsection{Gradient flows and dissipation}\label{sec:prelim-flows}

\begin{definition}[Gradient flow]\label{def:prelim-gradient-flow}
Fix $\gamma>0$. Let $(M,g)$ be a Riemannian manifold,
$I\subset\R$ a time interval, and $F:M\times I\to\R$ a function that is
$C^1$ jointly in $(p,t)$.
For each $t\in I$, define $F_t:M\to\R$ by $F_t(p):=F(p,t)$.
The gradient flow of $F$ with mobility $\gamma$ is the differential equation
\begin{equation}
\dot p(t)=-\gamma\grad_g F_t(p(t)).
\label{eq:prelim-gradient-flow}
\end{equation}
An initial-value problem also prescribes $p(t_0)=p_0\in M$.
\end{definition}

\begin{proposition}[Energy balance along a gradient flow]
\label{prop:prelim-dissipation}
Every solution of \eqref{eq:prelim-gradient-flow}
satisfies the following identity on its interval of existence:
\begin{equation*}
\frac{d}{dt}F(p(t),t)
=\partial_t F(p(t),t)
-\gamma\|\grad_g F_t(p(t))\|_{g,p(t)}^2.
\end{equation*}
In particular, if $F$ is time independent, it is nonincreasing along the flow.
\end{proposition}
\begin{proof}
The chain rule, \eqref{eq:prelim-gradient-flow} and
Definition~\ref{def:prelim-gradient} give
\[
\begin{aligned}
\frac{d}{dt}F(p(t),t)
&=\partial_t F(p(t),t)+d(F_t)_{p(t)}(\dot p(t))\\
&=\partial_t F(p(t),t)
-\gamma g_{p(t)}\bigl(\grad_g F_t(p(t)),\grad_g F_t(p(t))\bigr).
\end{aligned}
\]
\end{proof}

\subsection{Projection and constrained
gradients}\label{sec:prelim-projection-general}
\begin{figure}[ht]
\centering
\begin{tikzpicture}[scale=1.05,>=stealth]
\draw[->] (-0.8,0)--(4.2,0) node[right] {$V$};
\draw[->,thick] (0,0)--(2.5,1.6) node[above] {$v$};
\draw[->,thick] (0,0)--(2.5,0) node[below] {$\Pi v$};
\draw[->,dashed] (2.5,0)--(2.5,1.6) node[midway,right] {$v-\Pi v$};
\draw (2.3,0)--(2.3,0.2)--(2.5,0.2);
\node[below] at (0,0) {$0$};
\end{tikzpicture}
\caption{Projection in coordinates orthonormal for $g$.
The projected vector $\Pi v$ lies in $V$, and the
remainder $v-\Pi v$ is $g$-orthogonal to $V$.}
\label{fig:projection}
\end{figure}
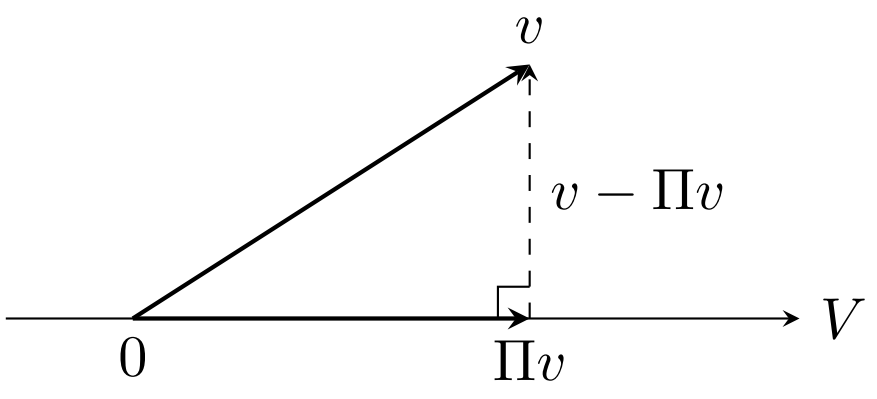

The orthogonal projection in Euclidean space extends to
Riemannian manifolds by using the inner products on their tangent spaces.
\begin{definition}[Orthogonal projection]\label{def:orthogonal-projection}
Let $(E,g)$ be a finite-dimensional inner-product space and $V\subset E$
a linear subspace. The $g$-orthogonal projection
$\Pi:E\to V$ is determined by
\[
\Pi v\in V,\qquad g(v-\Pi v,w)=0\quad\text{for every }v\in E
\text{ and }w\in V.
\]
\end{definition}
The orthogonal projection fixes vectors in $V$,
satisfies $\Pi^2=\Pi$, and is self-adjoint.
These properties and existence follow from the orthogonal decomposition
$E=V\oplus V^\perp$.
Figure~\ref{fig:projection} illustrates this decomposition.

\begin{definition}[Induced metric]\label{def:prelim-induced-metric}
Let $M$ be a smooth embedded submanifold of a Riemannian manifold
$(\mathcal N,h)$, with inclusion $\iota:M\hookrightarrow\mathcal N$.
The induced metric is the pullback $g=\iota^*h$, defined by
\[
g_p(v,w)=h_{\iota(p)}\bigl(d\iota_p(v),d\iota_p(w)\bigr),
\qquad p\in M,\quad v,w\in T_pM.
\]
\end{definition}
Identifying $T_pM$ with its image under $d\iota_p$,
Definition~\ref{def:orthogonal-projection} applies with
$E=T_p\mathcal N$, $V=T_pM$ and inner product $h_p$.
From now on, the word ``ambient'' will refer to the containing manifold $\mathcal N$.
\begin{proposition}[Gradient on an embedded submanifold]
\label{prop:prelim-submanifold-gradient}
Let $(\mathcal N,h)$ be a Riemannian manifold, and let
$M\subset\mathcal N$ be a smooth embedded submanifold equipped with the
metric $g$ induced by $h$. Let $\overline F:\mathcal N\to\R$ be $C^1$
and set $F=\overline F|_M$. For each $p\in M$, let
$\Pi_p:T_p\mathcal N\to T_pM$ be the $h_p$-orthogonal projection. Then
\begin{equation*}
\grad_g F(p)=\Pi_p\bigl(\grad_h\overline F(p)\bigr).
\end{equation*}
\end{proposition}
See Appendix~\ref{app:projection} for the proof.

\subsection{Probability vectors and the Fisher--Rao metric}
\label{sec:prelim-simplex}

Fix an integer $n\geq2$. Denote the outcome set by
$\Omega_n=\{1,\ldots,n\}$.

\begin{definition}[Open probability simplex]\label{def:prelim-simplex}
The space of strictly positive probability vectors is
\begin{equation*}
\mathcal P_n
=\left\{p\in(0,\infty)^n:\sum_{j=1}^n p_j=1\right\}.
\end{equation*}
\end{definition}
We identify $p\in\mathcal P_n$ with the probability
measure on $(\Omega_n,2^{\Omega_n})$ assigning mass $p_j$ to each outcome $j$.

\begin{lemma}[Dimension and tangent space]\label{lem:prelim-tangent}
The simplex $\mathcal P_n$ is a smooth embedded submanifold of
$U_n=(0,\infty)^n$ of dimension $n-1$. At each $p\in\mathcal P_n$,
we have
\begin{equation}
T_p\mathcal P_n
=\left\{v\in\R^n:\sum_{j=1}^n v_j=0\right\}.
\label{eq:prelim-tangent}
\end{equation}
\end{lemma}
See Appendix~\ref{app:simplex} for the proof.

\begin{definition}[Expectation]\label{def:prelim-expectation}
Let $V$ be a finite-dimensional real vector space. For
$p\in\mathcal P_n$ and a tuple
$b=(b_1,\ldots,b_n)\in V^n$, define its expectation by
\begin{equation*}
\E_p[b]=\sum_{j=1}^n p_jb_j\in V.
\end{equation*}
\end{definition}

For a fixed outcome $j\in\Omega_n$, define the log-probability
function
\[
\ell_j:\mathcal P_n\longrightarrow\R,
\qquad \ell_j(p)=\log p_j.
\]
At $p\in\mathcal P_n$, its differential is the linear map
\[
d\ell_j|_p:T_p\mathcal P_n\longrightarrow\R,
\qquad v\longmapsto\frac{v_j}{p_j}.
\]

\begin{definition}[Fisher--Rao metric]\label{def:prelim-fisher}
For $p\in\mathcal P_n$ and $v,w\in T_p\mathcal P_n$, set
\begin{equation}
g_p(v,w)
=\E_p\bigl[(d\ell_j|_p(v)\,d\ell_j|_p(w))_{j=1}^n\bigr]
=\sum_{j=1}^n\frac{v_jw_j}{p_j}.
\label{eq:prelim-fisher}
\end{equation}
\end{definition}
For each $p\in\mathcal P_n$, $g_p$ is bilinear and symmetric, and satisfies
\[
g_p(v,v)=\sum_{j=1}^n\frac{v_j^2}{p_j}>0
\qquad\text{for }v\ne0.
\]
Its coefficients depend smoothly on $p$, so $g$ defines a Riemannian
metric on $\mathcal P_n$. In this setting we also
write $\grad_{\FR}=\grad_g$ and $\|v\|_{\FR,p}^2=g_p(v,v)$.

\subsection{Computing the Fisher--Rao gradient}
\label{sec:prelim-projection}

We now apply the induced-metric construction to $\mathcal P_n\subset U_n$.
Equip the positive orthant with the ambient Riemannian metric
\begin{equation*}
\overline g_x(v,w)=v^\top D_xw,
\qquad D_x=\operatorname{diag}(x_1^{-1},\ldots,x_n^{-1}),
\end{equation*}
where $x\in U_n$ and $v,w\in T_xU_n\simeq\R^n$.
Its restriction to $T_p\mathcal P_n$ is exactly
\eqref{eq:prelim-fisher}. By
\eqref{eq:prelim-coordinate-gradient}, the inverse matrix
$D_p^{-1}=\operatorname{diag}(p_1,\ldots,p_n)$ converts the Euclidean
gradient on $U_n$ into the gradient for $\overline g$.

\begin{proposition}[Fisher--Rao gradient on the simplex]
\label{prop:prelim-gradient}
Let $F:\mathcal P_n\to\R$ be $C^1$, and choose a $C^1$
$\overline F:U_n\to\R$ with $\overline F|_{\mathcal P_n}=F$.
Define $f:\mathcal P_n\to\R^n$ by $f(p):=\nabla\overline F(p)$,
where $\nabla$ denotes the Euclidean gradient. For each
$p\in\mathcal P_n$,
\begin{equation*}
(\grad_{\FR}F(p))_j
=p_j\bigl(f_j(p)-\E_p[f(p)]\bigr)
=p_j\left(f_j(p)-\sum_{k=1}^n p_kf_k(p)\right),
\qquad j=1,\ldots,n.
\end{equation*}
Equivalently,
\begin{equation}
\grad_{\FR}F(p)=C(p)f(p),
\qquad C(p):=\operatorname{diag}(p)-pp^\top.
\label{eq:prelim-cometric}
\end{equation}
These formulas are independent of the ambient extension.
\end{proposition}
\begin{proof}
Such an extension exists: for example,
$\overline F(x)=F(x/\sum_{k=1}^n x_k)$ is $C^1$ on $U_n$ and agrees
with $F$ on $\mathcal P_n$.
Let $\Pi_p:T_pU_n\to T_p\mathcal P_n$ be the
$\overline g_p$-orthogonal projection, with $T_pU_n\simeq\R^n$.
The vector $p$ is an ambient unit normal to the simplex, since
\[
\overline g_p(p,v)=\sum_{j=1}^n v_j=0
\quad(v\in T_p\mathcal P_n),
\qquad
\overline g_p(p,p)=\sum_{j=1}^n p_j=1.
\]
Subtracting the normal component gives, for $z\in T_pU_n$,
\[
\Pi_pz=z-\overline g_p(z,p)p
=z-\left(\sum_{k=1}^n z_k\right)p.
\]
The ambient coordinate formula \eqref{eq:prelim-coordinate-gradient} gives
$\grad_{\overline g}\overline F(p)=D_p^{-1}f(p)$.
Applying Proposition~\ref{prop:prelim-submanifold-gradient},
\[
\grad_{\FR}F(p)
=\Pi_pD_p^{-1}f(p)
=\operatorname{diag}(p)f(p)-p\bigl(p^\top f(p)\bigr)=C(p)f(p).
\]
Independence of the extension also follows from this proposition.
\end{proof}

\subsection{Log-odds coordinates and the softmax Jacobian}
\label{sec:prelim-logits}

We now introduce global log-odds coordinates on the
$(n-1)$-dimensional manifold $\mathcal P_n$; see
\citet[Section 4.1, equations (47)--(49)]{NielsenSunMarchand2017}.

\begin{proposition}[Log-odds coordinates]\label{prop:prelim-log-odds}
With outcome $n$ as reference, the map
\[
\eta:\mathcal P_n\longrightarrow\R^{n-1},
\qquad \eta_j(p)=\log\frac{p_j}{p_n},
\quad j=1,\ldots,n-1,
\]
is a diffeomorphism. Its inverse is
\begin{equation}
p_n=\frac{1}{1+\sum_{k=1}^{n-1}e^{\eta_k}},
\qquad
p_j=\frac{e^{\eta_j}}{1+\sum_{k=1}^{n-1}e^{\eta_k}}
\quad(j=1,\ldots,n-1).
\label{eq:prelim-log-odds}
\end{equation}
\end{proposition}
\begin{proof}
The relations $p_j=e^{\eta_j}p_n$, for $j=1,\ldots,n-1$, and
$\sum_{k=1}^n p_k=1$ give \eqref{eq:prelim-log-odds}. The forward and inverse
maps are smooth.
\end{proof}

In affine simplex coordinates, negative entropy
$\mathcal E(p)=\sum_{j=1}^n p_j\log p_j$ has the log-odds as its
first derivatives and the Fisher--Rao metric matrix as its Hessian.
The calculation appears in Appendix~\ref{app:entropy-coordinates}.

Log-odds use $n-1$ coordinates relative to a reference outcome.
Softmax gives a representation treating all $n$ outcomes symmetrically.

\begin{definition}[Softmax and logits]\label{def:prelim-softmax}
The softmax map $\softmax:\R^n\to\mathcal P_n$ is defined by
\begin{equation*}
\softmax_j(u)=\frac{e^{u_j}}{\sum_{k=1}^n e^{u_k}},
\qquad j=1,\ldots,n.
\end{equation*}
The components of $u$ are called logits.
\end{definition}

\begin{proposition}[Logit representation]\label{prop:prelim-logit-representation}
The softmax map is surjective. For every $p\in\mathcal P_n$,
\[
\softmax^{-1}(p)=\{\log p+c\mathbf1:c\in\R\},
\qquad \mathbf1=(1,\ldots,1)^\top,
\]
where the logarithm is componentwise. If $p=\softmax(u)$, then
\[
\log\frac{p_j}{p_k}=u_j-u_k,\qquad j,k=1,\ldots,n.
\]
Consequently the logits form a redundant representation; imposing $u_n=0$
recovers the log-odds coordinates $u_j=\eta_j(p)$ for $j=1,\ldots,n-1$.
\end{proposition}
\begin{proof}
Since $\sum_{j=1}^n p_j=1$, we have $\softmax(\log p)=p$.
If $p=\softmax(u)$, taking logarithms gives
$u_j=\log p_j+\log\sum_{k=1}^n e^{u_k}$, so every representative has the
stated form. Conversely, adding a common constant to all logits leaves
softmax unchanged. Dividing $p_j$ by $p_k$, for
$j,k\in\{1,\ldots,n\}$, gives the log-odds identity, and setting $k=n$
proves the last assertion.
\end{proof}
For a $C^1$ scalar function $a(t)$, the shift
$u(t)\mapsto u(t)+a(t)\mathbf1$ leaves the probability path unchanged
and adds $\dot a(t)\mathbf1$ to the logit velocity. This explains the scalar freedom in
Lemma~\ref{lem:prelim-logit-flow} below.

The softmax derivative acts on vectors in $\R^n$, whereas the metric
map $g_p^\sharp$ acts on covectors (Definition~\ref{def:musical}).
To compare them, for $p\in\mathcal P_n$ and $f\in\R^n$, write
\[
[f]:T_p\mathcal P_n\longrightarrow\R,
\qquad [f](v):=f^\top v.
\]

\begin{proposition}[Softmax Jacobian and the metric isomorphism]
\label{prop:prelim-softmax-cometric}
For the softmax map $\softmax:\R^n\to\mathcal P_n$,
fix a logit vector $u\in\R^n$ and set $p=\softmax(u)$.
Then its Jacobian has entries
\begin{equation*}
\frac{\partial\softmax_j}{\partial u_k}(u)
=p_j(\delta_{jk}-p_k),\qquad j,k=1,\ldots,n.
\end{equation*}
Equivalently, the derivative $D\softmax(u):\R^n\to T_p\mathcal P_n$
is represented by the matrix $C(p)$ in \eqref{eq:prelim-cometric}.
Every covector in $T_p^*\mathcal P_n$ has the form
$[f]$, with $f$ determined up to addition of a multiple of $\mathbf1$.
In this notation, the inverse of $g_p^\flat$ is
\begin{equation*}
g_p^\sharp:T_p^*\mathcal P_n\longrightarrow T_p\mathcal P_n,
\qquad [f]\longmapsto C(p)f.
\end{equation*}
Thus the softmax Jacobian and the map taking covectors to Fisher
gradients are represented by the same matrix $C(p)$.
\end{proposition}
\begin{proof}
Differentiating the quotient gives
\[
\frac{\partial\softmax_j}{\partial u_k}(u)
=\frac{\delta_{jk}e^{u_j}\sum_{\ell=1}^n e^{u_\ell}
-e^{u_j}e^{u_k}}{(\sum_{\ell=1}^n e^{u_\ell})^2}
=p_j(\delta_{jk}-p_k).
\]
The Euclidean inner product on $T_p\mathcal P_n$ represents each
covector $\alpha\in T_p^*\mathcal P_n$ by a vector
$f\in T_p\mathcal P_n\subset\R^n$ satisfying
\[
\alpha(v)=f^\top v\qquad\text{for every }v\in T_p\mathcal P_n.
\]
Two vectors in $\R^n$ represent the same covector exactly when their
difference is orthogonal to $T_p\mathcal P_n$ in the Euclidean inner product.
By \eqref{eq:prelim-tangent}, this means that their difference is a
multiple of $\mathbf1$.

Since $C(p)\mathbf1=0$ and $\mathbf1^\top C(p)=0$, the assignment
$[f]\mapsto C(p)f$ is independent of the representative and takes
values in $T_p\mathcal P_n$.
For every $v\in T_p\mathcal P_n$,
\[
g_p(C(p)f,v)
=\sum_{j=1}^n\bigl(f_j-\E_p[f]\bigr)v_j=f^\top v.
\]
This is the defining identity of $g_p^\sharp$.
\end{proof}

\begin{lemma}[Lifting Fisher--Rao flow to logits]\label{lem:prelim-logit-flow}
Fix $\gamma>0$. Let $I\subset\R$ be a time interval and
$F:\mathcal P_n\times I\to\R$ a jointly $C^1$ function.
Choose a jointly $C^1$ extension $\overline F:U_n\times I\to\R$ and
define $f:\mathcal P_n\times I\to\R^n$ by
$f(p,t):=\nabla_p\overline F(p,t)$.
A $C^1$ logit path $u:I\to\R^n$ projects by
$p(t)=\softmax(u(t))$ to the Fisher--Rao gradient flow of $F$
(see Definition~\ref{def:prelim-gradient-flow}) if and only if,
for some continuous $c:I\to\R$,
\begin{equation*}
\dot u(t)=-\gamma f(p(t),t)+c(t)\mathbf1.
\end{equation*}
Every positive gradient-flow solution has such a lift.
\end{lemma}
\begin{proof}
Differentiating $p(t)=\softmax(u(t))$ gives, by
Proposition~\ref{prop:prelim-softmax-cometric},
\[
\dot p(t)=C(p(t))\dot u(t).
\]
Since $f$ is the Euclidean gradient of the extension,
$d(F_t)_{p(t)}(v)=f(p(t),t)^\top v$ for every
$v\in T_{p(t)}\mathcal P_n$. Hence $d(F_t)_{p(t)}=[f(p(t),t)]$.
Definition~\ref{def:prelim-gradient} and
Proposition~\ref{prop:prelim-softmax-cometric} therefore give
\[
\grad_{\FR}F_t(p(t))
=g_{p(t)}^\sharp\bigl(d(F_t)_{p(t)}\bigr)
=g_{p(t)}^\sharp\bigl([f(p(t),t)]\bigr)
=C(p(t))f(p(t),t).
\]
For $p\in\mathcal P_n$ and $z\in\R^n$,
$(C(p)z)_j=p_j(z_j-\E_p[z])$.
Since $p_j>0$, $C(p)z=0$ if and only if $z_j=\E_p[z]$ for
every $j=1,\ldots,n$, so $\ker C(p)=\operatorname{span}\{\mathbf1\}$.
Therefore,
\[
\begin{aligned}
\dot p(t)=-\gamma\grad_{\FR}F_t(p(t))
&\iff C(p(t))\bigl(\dot u(t)+\gamma f(p(t),t)\bigr)=0\\
&\iff \dot u(t)+\gamma f(p(t),t)\in\operatorname{span}\{\mathbf1\}\\
&\iff \dot u(t)=-\gamma f(p(t),t)+c(t)\mathbf1.
\end{aligned}
\]
Here $c(t)=\dot u_1(t)+\gamma f_1(p(t),t)$ is continuous
because $u$ is $C^1$ and $f$ is continuous.
Finally, every positive gradient-flow solution $p$ has the $C^1$ lift
$u(t)=\log p(t)$, since $\softmax(\log p(t))=p(t)$.
\end{proof}

\subsection{From one row to the attention matrix}
The row representation $A\leftrightarrow(a^{(1)},\ldots,a^{(N)})$
identifies $\mathcal A$ with $(\mathcal P_N)^N$.
\begin{definition}[Product metric]\label{def:prelim-product-metric}
For Riemannian manifolds $(M_r,g^{(r)})$, $r=1,\ldots,b$, the sum product
metric on $M_1\times\cdots\times M_b$ is
\begin{equation*}
g^{\mathrm{prod}}_p(U,V)
=\sum_{r=1}^b g^{(r)}_{p^{(r)}}\bigl(U^{(r)},V^{(r)}\bigr),
\qquad p=(p^{(1)},\ldots,p^{(b)}).
\end{equation*}
Here $U^{(r)},V^{(r)}\in T_{p^{(r)}}M_r$ are the factor components.
\end{definition}
For attention, the sum product metric extends the quadratic form
in \eqref{eq:free-energy-quadratic} to all $A\in\mathcal A$:
\begin{equation}\label{eq:attention-metric}
g_A(B,C)=\sum_{i,j=1}^N\frac{B_{ij}C_{ij}}{A_{ij}},\qquad
\|B\|_{\FR,A}^2=g_A(B,B),\qquad B,C\in T_A\mathcal A.
\end{equation}
Recall the row-expectation notation
$\E_i[b]=\sum_{j=1}^N A_{ij}b_j$ introduced in
Section~\ref{sec:independent-attention}.
Definition~\ref{def:prelim-expectation} identifies it with
$\E_{a^{(i)}}[b]$.
Along a solution, its weights are $A_{ij}(t)$.

\begin{proposition}[Geometry of the attention space]\label{prop:matrix-geometry}
The manifold $\mathcal A$ has dimension $N(N-1)$, and
\[
T_A\mathcal A=\left\{B\in\R^{N\times N}:
\sum_{j=1}^N B_{ij}=0\ \text{for }i=1,\ldots,N\right\}.
\]
For a smooth $F:\mathcal A\to\R$, choose a smooth extension $\bar F$
to the positive matrices. Set $f_{ij}=\partial\bar F/\partial A_{ij}$
and $f_i=(f_{i1},\ldots,f_{iN})^\top$. Then
\begin{equation}
(\grad_A^{\FR}F)_{ij}
=A_{ij}\bigl(f_{ij}-\E_i[f_i]\bigr),
\qquad i,j=1,\ldots,N.
\label{eq:matrix-gradient}
\end{equation}
This expression is independent of the extension.
\end{proposition}
\begin{proof}
The dimension and tangent space follow by applying
Lemma~\ref{lem:prelim-tangent} to every factor. By the product-metric
construction, the $i$th row of the gradient is the simplex gradient
of $F$ with all other rows held fixed. Applying
Proposition~\ref{prop:prelim-gradient} to this function gives
\eqref{eq:matrix-gradient}, including independence of the extension.
\end{proof}

\section{Attention relaxation and tracking}\label{sec:attention-flow}
\subsection{The evolution law}
During a Transformer forward pass, successive layers update the token
representations and compute attention from the resulting features
\citep{Vaswani2017}. We idealise such changes by a prescribed continuous
path $X:[0,\infty)\to\mathcal X$;
Appendix~\ref{sec:depth} discusses the connection with depth.
Standard self-attention computes its associated weights directly by softmax.
Giving attention its own evolution is the extension studied here:
varying $X(t)$ lets us study how closely this evolution tracks the
changing softmax target $A^*(X(t))$.

We choose the Fisher--Rao gradient
flow of \eqref{eq:total-free-energy} as an evolution law for $A$, using
the metric \eqref{eq:attention-metric} on $\mathcal A$
(see Definition~\ref{def:prelim-gradient-flow}).
This describes continuous-time steepest descent of the attention
free energy.
Section~\ref{sec:hamiltonian} will determine $X(t)$ as part of a coupled
initial-value problem.

\begin{definition}[Fisher--Rao attention model]\label{def:attention-model}
Let $X:[0,\infty)\to\mathcal X$ be a prescribed $C^1$ feature path.
Fix a mobility $\gamma>0$. For $A^0\in\mathcal A$, we model
the attention weights by the Fisher--Rao gradient flow
\begin{equation}
\frac{dA}{dt}=-\gamma\grad_A^{\FR}\cF(A(t);X(t)),
\qquad A(0)=A^0.
\label{eq:attention-flow}
\end{equation}
Here $\cF$ is the free energy defined in
\eqref{eq:total-free-energy}.
\end{definition}

\begin{proposition}[The attention gradient]\label{prop:attention-gradient}
For every $A\in\mathcal A$ and $X\in\mathcal X$,
and for $i,j=1,\ldots,N$,
\begin{equation*}
(\grad_A^{\FR}\cF)_{ij}
=A_{ij}\left[-s_{ij}(X)+\E_i[s_i(X)]
+\tau\bigl(\log A_{ij}-\E_i[\log a^{(i)}]\bigr)\right].
\end{equation*}
Consequently \eqref{eq:attention-flow} has the component form
\begin{equation}
\dot A_{ij}=\gamma A_{ij}\bigl[s_{ij}(X(t))-\E_i[s_i(X(t))]
-\tau\bigl(\log A_{ij}-\E_i[\log a^{(i)}]\bigr)\bigr].
\label{eq:attention-flow1}
\end{equation}
\end{proposition}
\begin{proof}
For the extension of \eqref{eq:total-free-energy} to positive matrices,
\begin{equation}
f_{ij}:=\frac{\partial\bar{\cF}}{\partial A_{ij}}
=-s_{ij}(X)+\tau(1+\log A_{ij}).
\label{eq:energy-derivatives}
\end{equation}
Its row mean is
\[
\sum_{k=1}^N A_{ik}f_{ik}
=-\sum_{k=1}^N A_{ik}s_{ik}(X)
+\tau\left(1+\sum_{k=1}^N A_{ik}\log A_{ik}\right).
\]
Subtracting this mean from $f_{ij}$ cancels the constant $\tau$.
Formula~\eqref{eq:matrix-gradient} then gives the result.
\end{proof}

\begin{remark}[Replicator form]\label{rem:attention-replicator}
With $r_{ij}(A,X)=s_{ij}(X)-\tau\log A_{ij}$ and
$r_i=(r_{i1},\ldots,r_{iN})^\top$, equation~\eqref{eq:attention-flow1} becomes
\[
\dot A_{ij}=\gamma A_{ij}\bigl(r_{ij}-\E_i[r_i]\bigr).
\]
Each row therefore has the replicator form, with the state-dependent
payoff $r_{ij}$ containing the entropy contribution.
For the connection between entropy-adjusted replicator
dynamics and discounted scores, see \citet{Coucheney2015}.
\end{remark}

\subsection{Logit and log-odds formulations}
The row softmax map
$\softmax_{\mathrm{row}}:\R^{N\times N}\to\mathcal A$ applies softmax
to each row. For a logit matrix $U\in\R^{N\times N}$, write
\begin{equation*}
A=\softmax_{\mathrm{row}}(U),\qquad
A_{ij}=\frac{e^{U_{ij}}}{\sum_{k=1}^N e^{U_{ik}}},
\qquad i,j=1,\ldots,N.
\end{equation*}
\begin{proposition}[Row-wise logit lift]\label{prop:attention-logit-lift}
Let $X$ and $A^0$ be as in Definition~\ref{def:attention-model}.
Let $U:[0,\infty)\to\R^{N\times N}$ solve
\begin{equation}\label{eq:logit-flow}
\dot U=\gamma(S(X(t))-\tau U),\qquad U(0)=\log A^0.
\end{equation}
Then $A=\softmax_{\mathrm{row}}(U)$ solves
\eqref{eq:attention-flow}. For prescribed $X$, equation~\eqref{eq:logit-flow}
is linear in $U$.
\end{proposition}
\begin{proof}
Since $\log A_{ij}=U_{ij}-\log\sum_{k=1}^N e^{U_{ik}}$, the derivatives
in \eqref{eq:energy-derivatives} have the form
\[
f_{ij}=-s_{ij}(X(t))+\tau U_{ij}
+\tau\left(1-\log\sum_{k=1}^N e^{U_{ik}}\right),
\qquad i,j=1,\ldots,N.
\]
Apply Lemma~\ref{lem:prelim-logit-flow} row by row with
\[
c_i(t)=\gamma\tau\left(1-\log\sum_{k=1}^N e^{U_{ik}(t)}\right),
\qquad i=1,\ldots,N.
\]
With this choice, we obtain
\[
-\gamma f_{ij}+c_i(t)=\gamma(s_{ij}(X(t))-\tau U_{ij}),
\]
which gives \eqref{eq:logit-flow}.
Normalisation gives $\softmax_{\mathrm{row}}(\log A^0)=A^0$.
\end{proof}

For the uniqueness and tracking arguments, it is useful to pass to
log-odds, which remove the arbitrary row shifts in $U$. Using
Proposition~\ref{prop:prelim-log-odds}, set
\[
\eta_{ij}=\log\frac{A_{ij}}{A_{iN}}=U_{ij}-U_{iN},
\qquad i=1,\ldots,N,\quad j=1,\ldots,N-1.
\]
Differentiating these log-odds using \eqref{eq:attention-flow1},
or subtracting the lifted equations for $U_{ij}$ and $U_{iN}$, gives
\begin{equation}\label{eq:log-odds-flow}
\dot\eta_{ij}=\gamma\bigl(s_{ij}(X(t))-s_{iN}(X(t))-\tau\eta_{ij}\bigr),
\qquad i=1,\ldots,N,\quad j=1,\ldots,N-1.
\end{equation}

\subsection{Existence and filtered scores}

\begin{theorem}[Attention driven by a prescribed feature path]
\label{thm:attention-solution}
Let $X:[0,\infty)\to\mathcal X$ be a prescribed $C^1$ feature path.
For every $A^0\in\mathcal A$, the initial-value problem
\eqref{eq:attention-flow} has a unique global solution
$A:[0,\infty)\to\mathcal A$.
With $\kappa=\gamma\tau$, define $U:[0,\infty)\to\R^{N\times N}$ by
the first line below. The attention solution is its row softmax:
\begin{equation}
\begin{aligned}
U(t)&=e^{-\kappa t}\log A^0
+\gamma\int_0^t e^{-\kappa(t-r)}S(X(r))\,dr,\\
A(t)&=\softmax_{\mathrm{row}}(U(t)).
\end{aligned}
\label{eq:attention-solution}
\end{equation}
In particular, all entries stay strictly positive and each row remains
normalised at every finite time.
\end{theorem}
\begin{proof}
Multiplying \eqref{eq:logit-flow} by $e^{\kappa t}$ gives
\[
\frac{d}{dt}\bigl(e^{\kappa t}U(t)\bigr)
=\gamma e^{\kappa t}S(X(t)).
\]
Integration gives \eqref{eq:attention-solution}. The score matrix is
continuous and hence bounded on every finite time interval, so every
entry of $U(t)$ is finite there. Its row softmax is positive and
normalised. At time zero,
$\softmax_{\mathrm{row}}(\log A^0)=A^0$;
Proposition~\ref{prop:attention-logit-lift}
verifies the differential equation.

For uniqueness, any solution in $\mathcal A$ has log-odds satisfying
\eqref{eq:log-odds-flow}, with initial values
$\eta_{ij}(0)=\log(A^0_{ij}/A^0_{iN})$
for $i=1,\ldots,N$ and $j=1,\ldots,N-1$. Uniqueness
for these scalar linear equations fixes all ratios $A_{ij}/A_{iN}$.
Normalisation fixes each remaining entry $A_{iN}$. Thus every such
solution agrees with the one constructed above, which exists for all
$t\geq0$.
\end{proof}

\begin{remark}[Exponential filtering]
The first line of \eqref{eq:attention-solution} can be written as
\[
U(t)=e^{-\kappa t}\log A^0
+\int_0^t\kappa e^{-\kappa(t-r)}\frac{S(X(r))}{\tau}\,dr.
\]
The initial logits and the past target logits $S(X(r))/\tau$ carry
weights whose sum is $e^{-\kappa t}+1-e^{-\kappa t}=1$.
Thus this model computes the softmax of exponentially filtered
scores,
with memory time $1/\kappa=1/(\gamma\tau)$. The averaging takes place
in logit space before the softmax is applied.
\end{remark}

\subsection{Tracking changing scores}
We now compare the attention solution of Theorem~\ref{thm:attention-solution}
with $A^*(X(t))$, the softmax associated with the current features.
For each receiving row $i=1,\ldots,N$, use source $N$ as reference.
The target and evolving log-odds are
\[
\begin{aligned}
h_{ij}(t)&:=\log\frac{A^*_{ij}(X(t))}{A^*_{iN}(X(t))}
=\frac{s_{ij}(X(t))-s_{iN}(X(t))}{\tau},\\
\eta_{ij}(t)&=\log\frac{A_{ij}(t)}{A_{iN}(t)},
\qquad j=1,\ldots,N-1.
\end{aligned}
\]
Thus $h,\eta:[0,\infty)\to\R^{N\times(N-1)}$, and
$E:=\eta-h$ measures the tracking error in log-odds.
For $E\in\R^{N\times(N-1)}$ and $B\in\R^{r\times s}$, write
\[
\|E\|_\infty:=\max_{\substack{1\leq i\leq N\\1\leq j\leq N-1}}|E_{ij}|,
\qquad
\|B\|_{\mathrm F}:=\left(\sum_{i=1}^r\sum_{j=1}^s B_{ij}^2\right)^{1/2}.
\]
We also denote the operator norm induced by the Euclidean
norms by $\|B\|_{\mathrm{op}}:=\sup_{v\in\R^s,\,|v|=1}|Bv|$.

\begin{theorem}[Tracking and convergence for changing features]
\label{thm:tracking}
Let $X,A$ be as in Theorem~\ref{thm:attention-solution}, and let $h,\eta,E$
be the paths defined above. Put $\kappa=\gamma\tau$.
For every $t\geq0$, the log-odds error satisfies
\begin{equation}\label{eq:tracking-bound}
\begin{aligned}
\dot E+\kappa E&=-\dot h,\\
\|E(t)\|_\infty
&\leq e^{-\kappa t}\|E(0)\|_\infty
+\int_0^te^{-\kappa(t-r)}\|\dot h(r)\|_\infty\,dr.
\end{aligned}
\end{equation}
In particular, if $\|\dot h(t)\|_\infty\leq L$ for all $t\geq0$, then
\[
\|E(t)\|_\infty\leq e^{-\kappa t}\|E(0)\|_\infty
+\frac L\kappa(1-e^{-\kappa t}).
\]
There are two useful sufficient conditions for asymptotic agreement:
\begin{enumerate}
\item If $\dot h(t)\to0$ as $t\to\infty$, then $E(t)\to0$ and
$\|A(t)-A^*(X(t))\|_{\mathrm F}\to0$.
\item If $h(t)\to h^\infty\in\R^{N\times(N-1)}$, then
$A(t)$ and $A^*(X(t))$ converge to the same limit
$A^\infty\in\mathcal A$, given row by row by
\[
A^\infty_{iN}=\frac1{1+\sum_{k=1}^{N-1}e^{h^\infty_{ik}}},
\qquad
A^\infty_{ij}=\frac{e^{h^\infty_{ij}}}
{1+\sum_{k=1}^{N-1}e^{h^\infty_{ik}}},
\]
for $i=1,\ldots,N$ and $j=1,\ldots,N-1$.
\end{enumerate}
The second condition holds in particular if $X(t)\to X^\infty$,
in which case $A^\infty=A^*(X^\infty)$.
\end{theorem}
\begin{proof}
Equation~\eqref{eq:log-odds-flow} is
$\dot\eta(t)=\kappa(h(t)-\eta(t))$. Since $E=\eta-h$,
subtracting $\dot h(t)$ gives $\dot E(t)+\kappa E(t)=-\dot h(t)$.
Multiplying by the integrating factor $e^{\kappa t}$ yields
\[
\frac{d}{dt}\bigl(e^{\kappa t}E(t)\bigr)
=e^{\kappa t}\bigl(\dot E(t)+\kappa E(t)\bigr)
=-e^{\kappa t}\dot h(t).
\]
Integrating from $0$ to $t$ and multiplying by $e^{-\kappa t}$ gives
\[
E(t)=e^{-\kappa t}E(0)
-\int_0^te^{-\kappa(t-r)}\dot h(r)\,dr.
\]
Taking the maximum norm proves the estimates. If $\dot h(t)\to0$,
split the integral at a fixed $T$. Its part on $[0,T]$ decays
exponentially, while its remaining norm is bounded by
$\sup_{r\geq T}\|\dot h(r)\|_\infty/\kappa$. Letting first $t$ and
then $T$ tend to infinity gives $E(t)\to0$.
To pass from log-odds to probabilities, use the Jacobian in
Proposition~\ref{prop:prelim-softmax-cometric}. The symmetric matrix
$C(p)$ has absolute row sums $2p_j(1-p_j)\leq1/2$, so
$\|C(p)\|_{\mathrm{op}}\leq1/2$ for every $p\in\mathcal P_N$.
Applying the mean-value estimate to each pair
$(\eta_{i1},\ldots,\eta_{i,N-1},0)$ and
$(h_{i1},\ldots,h_{i,N-1},0)$ gives
\[
\|A(t)-A^*(X(t))\|_{\mathrm F}
\leq\tfrac12\|E(t)\|_{\mathrm F}
\leq\tfrac12\sqrt{N(N-1)}\,\|E(t)\|_\infty.
\]
This proves the matrix tracking assertion.

For the second condition, use the equation for $\eta$ directly:
\[
\eta(t)-h^\infty=e^{-\kappa t}(\eta(0)-h^\infty)
+\kappa\int_0^te^{-\kappa(t-r)}(h(r)-h^\infty)\,dr.
\]
The same splitting argument gives $\eta(t)\to h^\infty$.
The inverse log-odds map in Proposition~\ref{prop:prelim-log-odds}
then gives the displayed limit for $A$;
applying it to $h(t)$ gives the same limit for $A^*(X(t))$.
Continuity of the bilinear scores proves the assertion for convergent
features.
\end{proof}
The first condition permits a slowly moving target with no limit.
The second gives convergence to a fixed attention matrix without
requiring the feature vectors themselves to settle. Persistent score
variation can instead leave a nonzero lag.

\begin{corollary}[Convergence to softmax at fixed features]
\label{thm:fixed-score-convergence}
Fix $X\in\mathcal X$.
The unique equilibrium of \eqref{eq:attention-flow} in $\mathcal A$
is $A^*(X)$. For every $A^0\in\mathcal A$, its solution satisfies
\[
\lim_{t\to\infty}\|A(t)-A^*(X)\|_{\mathrm F}=0.
\]
More precisely, for all $i,j,k\in\{1,\ldots,N\}$,
\begin{equation}
\log\frac{A_{ij}(t)}{A_{ik}(t)}
-\log\frac{A^*_{ij}(X)}{A^*_{ik}(X)}
=e^{-\gamma\tau t}\left(
\log\frac{A^0_{ij}}{A^0_{ik}}
-\log\frac{A^*_{ij}(X)}{A^*_{ik}(X)}\right).
\label{eq:log-ratio-convergence}
\end{equation}
\end{corollary}
\begin{proof}
For constant scores, $\dot h=0$ and \eqref{eq:tracking-bound} gives
$E(t)=e^{-\gamma\tau t}E(0)$. Theorem~\ref{thm:tracking} yields
matrix convergence. Subtracting two source log-odds gives
\eqref{eq:log-ratio-convergence} for arbitrary $j,k$.
At an equilibrium, \eqref{eq:log-odds-flow} forces the softmax ratios;
normalisation fixes $A=A^*(X)$, which is indeed stationary.
\end{proof}
\begin{example}[Uniform initial attention]
\label{ex:uniform-attention}
Fix $X\in\mathcal X$ and take $A^0_{ij}=1/N$ for $i,j=1,\ldots,N$.
Since $\log A^0_{ij}=-\log N$, Theorem~\ref{thm:attention-solution} gives
\[
U_{ij}(t)=-e^{-\gamma\tau t}\log N
+\bigl(1-e^{-\gamma\tau t}\bigr)\frac{s_{ij}(X)}{\tau},
\qquad i,j=1,\ldots,N.
\]
The first term is constant within each row and cancels under softmax.
Hence
\[
A(t)=\softmax_{\mathrm{row}}\left(
\bigl(1-e^{-\gamma\tau t}\bigr)\frac{S(X)}{\tau}\right).
\]
This formula exhibits the evolution from uniform attention to the
equilibrium weights.
\end{example}

\subsection{The attention energy balance}
We apply the gradient-flow identity in
Proposition~\ref{prop:prelim-dissipation} to the attention model
\eqref{eq:attention-flow}, using the gradient computed in
Proposition~\ref{prop:attention-gradient}.
For $i=1,\ldots,N$, write
$\dot s_i(t):=\frac{d}{dt}s_i(X(t))$ for the score derivative
along the prescribed feature path.

\begin{proposition}[Free-energy and relative-entropy balances]
\label{prop:attention-energy-balance}
Let $X:[0,\infty)\to\mathcal X$ be a prescribed $C^1$ feature path,
and let $A:[0,\infty)\to\mathcal A$ be the corresponding attention
trajectory from Theorem~\ref{thm:attention-solution}.
For every $t\geq0$,
\begin{equation}
\frac{d}{dt}\cF(A(t);X(t))
=-\sum_{i=1}^N\E_i[\dot s_i(t)]
-\gamma\|\grad_A^{\FR}\cF(A(t);X(t))\|_{\FR,A(t)}^2.
\label{eq:attention-energy-balance}
\end{equation}
If the feature path is constant, $X(t)\equiv X_0\in\mathcal X$, then
\begin{equation*}
\frac{d}{dt}\cF(A(t);X_0)
=-\gamma\|\grad_A^{\FR}\cF(A(t);X_0)\|_{\FR,A(t)}^2.
\end{equation*}
In particular, for the equilibrium rows $a^{*(i)}$ of $A^*(X_0)$,
the free-energy gap identity \eqref{eq:free-energy-gap} gives
\begin{equation}
\frac{d}{dt}\sum_{i=1}^N
\KL\bigl(a^{(i)}(t)\|a^{*(i)}\bigr)
=-\frac{\gamma}{\tau}
\|\grad_A^{\FR}\cF(A(t);X_0)\|_{\FR,A(t)}^2.
\label{eq:kl-dissipation}
\end{equation}
Both the free energy and this total relative entropy have strictly
negative derivatives whenever $A(t)\ne A^*(X_0)$.
\end{proposition}
\begin{proof}
Apply Proposition~\ref{prop:prelim-dissipation} to
$(A,t)\mapsto\cF(A;X(t))$ on $\mathcal A\times[0,\infty)$.
With $A$ held fixed, its time derivative is
\[
\partial_t\cF(A;X(t))
=-\sum_{i,j=1}^N A_{ij}\frac{d}{dt}s_{ij}(X(t)).
\]
Evaluating at $A=A(t)$ gives \eqref{eq:attention-energy-balance}.

To identify the dissipation, let $f_{ij}(t)$ be the derivatives
from \eqref{eq:energy-derivatives} evaluated at $(A(t);X(t))$,
and write $f_i(t)=(f_{i1}(t),\ldots,f_{iN}(t))^\top$.
Substitution of \eqref{eq:matrix-gradient} into the metric
\eqref{eq:attention-metric} gives
\begin{equation}
\begin{aligned}
\|\grad_A^{\FR}\cF(A(t);X(t))\|_{\FR,A(t)}^2
&=\sum_{i,j=1}^N
\frac{A_{ij}(t)^2\bigl(f_{ij}(t)-\E_i[f_i(t)]\bigr)^2}{A_{ij}(t)}\\
&=\sum_{i,j=1}^N
A_{ij}(t)\bigl(f_{ij}(t)-\E_i[f_i(t)]\bigr)^2\\
&=\sum_{i=1}^N\E_i\!\left[
\bigl(f_i(t)-\E_i[f_i(t)]\mathbf1\bigr)^2\right].
\end{aligned}
\label{eq:attention-variance}
\end{equation}
Here the square inside the expectation is taken componentwise.
Every $A_{ij}(t)>0$, so this norm vanishes precisely when
$f_{ij}(t)$ is constant in $j$ within each row.
By \eqref{eq:energy-derivatives} and row normalisation, this is
equivalent to $A(t)=A^*(X(t))$, as in the proof of
Proposition~\ref{prop:softmax-minimiser}.
When $X(t)\equiv X_0$, the scores are constant, so the first term in
\eqref{eq:attention-energy-balance} vanishes, giving the stated
free-energy dissipation identity. Differentiating
\eqref{eq:free-energy-gap} and using this identity gives
\eqref{eq:kl-dissipation}, since $\cF(A^*(X_0);X_0)$ is constant.
The equality condition above proves the strict decrease of both
quantities away from equilibrium.
\end{proof}
The derivatives $f_{ij}(t)$ are marginal free-energy costs. Attention
relaxes by shifting weight from above-average to below-average
costs, with averages taken under the current attention rows.
Equation~\eqref{eq:attention-variance} shows that the dissipation rate
is $\gamma$ times the sum of the corresponding row variances;
it vanishes exactly when $A(t)=A^*(X(t))$.
For fixed features, this decreases the total KL discrepancy from
softmax as well as the free energy.

When the features vary, changing the scores changes $\cF$ even with
the weights held fixed. This is the first term in
\eqref{eq:attention-energy-balance}; it can have either sign and
can outweigh the attention dissipation.

\section{Hamiltonian coupling}\label{sec:hamiltonian}
We now determine the joint evolution of $X(t)$ and $A(t)$.
The feature dynamics are Hamiltonian, and the attention relaxes by
Fisher--Rao descent of the same energy.

\subsection{The canonical model and its derivation}
We introduce momenta $P=(p_1,\ldots,p_N)\in\R^{d\times N}$,
identifying feature covectors with matrices via the Frobenius pairing.
We consider the state space
\[
\mathcal Z=T^*\mathcal X\times\mathcal A
\simeq\mathcal X\times\R^{d\times N}\times\mathcal A,
\qquad (X,P,A)\in\mathcal Z.
\]
For $m>0$, define the Hamiltonian $H:\mathcal Z\to\R$ by
\begin{equation}\label{eq:hamiltonian-energy}
H(X,P,A)=\frac{1}{2m}\lVert P\rVert_{\mathrm F}^2+\cF(A;X),
\end{equation}
where $\lVert P\rVert_{\mathrm F}^2=\sum_{i=1}^N\sum_{\alpha=1}^d
P_{\alpha i}^2$ is the squared Frobenius norm.
For smooth observables $f,g:\mathcal Z\to\R$, hold $A$ fixed
and define the canonical Poisson bracket in the feature and momentum variables
by
\[
\{f,g\}
=\langle\nabla_X f,\nabla_P g\rangle_{\mathrm F}
-\langle\nabla_P f,\nabla_X g\rangle_{\mathrm F}.
\]
We couple its Hamiltonian equations to
Fisher--Rao descent in $A$, using the metric \eqref{eq:attention-metric}.
The resulting system is
\begin{equation}\label{eq:canonical-coupled}
\begin{aligned}
\dot X&=\nabla_P H=P/m,\\
\dot P&=-\nabla_X H=-\nabla_X\cF(A;X),\\
\dot A&=-\gamma\grad_A^{\FR}H
       =-\gamma\grad_A^{\FR}\cF(A;X).
\end{aligned}
\end{equation}
All partial gradients hold the other state variables fixed.
Note that the attention equation is \eqref{eq:attention-flow}, since the
kinetic energy is independent of $A$.

For a smooth observable $f:\mathcal Z\to\R$, time dependence enters
through the solution $(X(t),P(t),A(t))$. The chain rule gives
\begin{equation}\label{eq:hamiltonian-observable}
\begin{aligned}
\frac{d}{dt}f(X(t),P(t),A(t))
&=d_Xf[\dot X]+d_Pf[\dot P]+d_Af[\dot A]\\
&=\{f,H\}
-\gamma\,g_A\bigl(\grad_A^{\FR}f,\grad_A^{\FR}H\bigr),
\end{aligned}
\end{equation}
with all terms evaluated along the trajectory.

We prescribe $X(0)=X^0$, $P(0)=P^0$, and $A(0)=A^0$, where
$X^0,P^0\in\R^{d\times N}$ and $A^0\in\mathcal A$. For input features
$X^0$, a simple choice is $P^0=0$ with either uniform attention
$A^0_{ij}=1/N$ for $i,j=1,\ldots,N$, or $A^0=A^*(X^0)$.

\begin{proposition}[Force and energy balance]\label{prop:coupled-force-energy}
For $A\in\mathcal A$ and $X\in\mathcal X$, the feature force in
\eqref{eq:canonical-coupled} is
\begin{equation}\label{eq:score-force}
-\nabla_X\cF=MXA^\top+M^\top XA.
\end{equation}
Along any solution of \eqref{eq:canonical-coupled},
\begin{equation}\label{eq:coupled-energy}
\frac{dH}{dt}
=-\gamma\lVert \grad_A^{\FR}\cF\rVert_{\FR,A}^2
=-\frac1\gamma g_A(\dot A,\dot A)\leq0.
\end{equation}
\end{proposition}
\begin{proof}
Hold $A$ fixed and write the free energy as
\[
\cF(A;X)=-\sum_{k,j=1}^N A_{kj}x_k^\top Mx_j
        +\tau\sum_{k,j=1}^N A_{kj}\log A_{kj}.
\]
The entropy term is independent of $X$. Differentiating the two
occurrences of the feature vectors in each score gives, for
$i=1,\ldots,N$,
\begin{align*}
-\nabla_{x_i}\cF
&=\sum_{k,j=1}^N A_{kj}
  \bigl(\delta_{ik}Mx_j+\delta_{ij}M^\top x_k\bigr)\\
&=\sum_{j=1}^N A_{ij}Mx_j+
  \sum_{k=1}^N A_{ki}M^\top x_k.
\end{align*}
These are the columns of \eqref{eq:score-force}.
For the energy, the chain rule underlying
\eqref{eq:hamiltonian-observable} gives
\[
\dot H=\langle P/m,\dot P\rangle_{\mathrm F}
       +\langle \nabla_X\cF,\dot X\rangle_{\mathrm F}
       +d_A\cF[\dot A].
\]
The first two terms cancel by Hamilton's equations; metric duality
and $\dot A=-\gamma\grad_A^{\FR}\cF$ give \eqref{eq:coupled-energy}.
\end{proof}

The two force terms arise because $x_i$ enters the scores both as a
query and as a key. Neither $M$ nor $A$ needs to be symmetric.

\begin{remark}[Lagrange--d'Alembert formulation]
In local coordinates $q=(q^1,\ldots,q^r)$, a Lagrangian $L(q,\dot q)$ and a
Rayleigh dissipation function $\mathcal R(q,\dot q)$ give
\[
\frac{d}{dt}\frac{\partial L}{\partial\dot q^\alpha}
-\frac{\partial L}{\partial q^\alpha}
=f_\alpha=-\frac{\partial\mathcal R}{\partial\dot q^\alpha},
\qquad \alpha=1,\ldots,r,
\]
where $f$ is the additional force covector and $q$ is held fixed
in the velocity derivative.

For the coupled model, take $q=(X,A)$ and choose
\[
L(X,A,\dot X,\dot A)=\frac m2\lVert \dot X\rVert_{\mathrm F}^2-\cF(A;X),
\qquad
\mathcal R(A,\dot A)=\frac{1}{2\gamma}
        g_A(\dot A,\dot A).
\]
This is the quadratic dissipation function associated with the
Fisher--Rao metric; see \citet[Example 2.4 and Section
5.1]{ZhuMielke2024}.
Only the features have inertia: $P=m\dot X$, while the momentum
conjugate to $A$ vanishes because $L$ is independent of $\dot A$.
Thus $L$ is degenerate.

Since $\mathcal R$ has no $\dot X$ dependence, the feature equation is
\[
\frac{d}{dt}\nabla_{\dot X}L-\nabla_X L
=-\nabla_{\dot X}\mathcal R=0.
\]
Substituting $\nabla_{\dot X}L=m\dot X$ and
$\nabla_X L=-\nabla_X\cF$ gives $m\ddot X=-\nabla_X\cF$.

For attention, the absence of $\dot A$ in $L$ gives
$d_A\cF=-d_{\dot A}\mathcal R$. At fixed $A$, for $V\in T_A\mathcal A$,
\[
d_{\dot A}\mathcal R[V]
=\frac{g_A(V,\dot A)+g_A(\dot A,V)}{2\gamma}
=\frac1\gamma g_A(\dot A,V).
\]
Hence $d_A\cF=-\gamma^{-1}g_A^\flat\dot A$.
Using $\grad_A^{\FR}\cF=g_A^\sharp d_A\cF$
(Definition~\ref{def:prelim-gradient}), we obtain
\[
\dot A=-\gamma\grad_A^{\FR}\cF,
\]
the third equation in \eqref{eq:canonical-coupled}.
\end{remark}

\begin{remark}[Additional feature potentials]\label{rem:feature-potentials}
For a smooth function $\Psi:\R^{d\times N}\to\R$, replacing $H$ by
$H+\Psi(X)$ leaves the attention equation unchanged and adds
$-\nabla_X\Psi$ to the feature force. The same cancellation proves
\eqref{eq:coupled-energy} with $H+\Psi$ as the total energy.
\end{remark}

\subsection{Connection with energy-based attention}
For each fixed $X\in\mathcal X$, Proposition~\ref{prop:softmax-minimiser}
gives
\begin{equation}\label{eq:reduced-potential}
V(X):=\min_{A\in\mathcal A}\cF(A;X)
=-\tau\sum_{i=1}^N\log\left(\sum_{j=1}^N
                       e^{s_{ij}(X)/\tau}\right).
\end{equation}
Combining \eqref{eq:hamiltonian-energy} with \eqref{eq:free-energy-gap}
gives the exact decomposition
\begin{equation}\label{eq:hamiltonian-gap}
H(X,P,A)=\frac{1}{2m}\lVert P\rVert_{\mathrm F}^2+V(X)
 +\tau\sum_{i=1}^N\KL\bigl(a^{(i)}\Vert a^{*(i)}(X)\bigr).
\end{equation}
Thus the excess over the equilibrated Hamiltonian is exactly
the free-energy gap in \eqref{eq:free-energy-gap}.

The reduced potential \eqref{eq:reduced-potential} has the same
log-sum-exp form as the attention energy in Energy Transformer
\citep{Hoover2023}. Our model adds feature inertia and gives $A$ its own
Fisher--Rao relaxation, rather than computing attention directly from
the current scores.

Note that computing attention directly from the current scores,
by imposing $A=A^*(X)$, defines a reduced version of our Hamiltonian
model with energy $\lVert P\rVert_{\mathrm F}^2/(2m)+V(X)$.
The envelope identity is
\[
\nabla_X V(X)=\left.\nabla_X\cF(A;X)\right|_{A=A^*(X)},
\]
where the gradient on the right differentiates $X$ with $A$ held fixed.
Indeed, $V(X)=\cF(A^*(X);X)$, so for a feature variation
$Z\in\R^{d\times N}$ the chain rule gives
\[
dV(X)[Z]=d_X\cF(A^*(X);X)[Z]
+d_A\cF(A^*(X);X)[\mathrm DA^*(X)[Z]].
\]
Differentiating row normalisation gives
$\sum_{j=1}^N\mathrm DA^*_{ij}(X)[Z]=0$ for each row $i$, so
$\mathrm DA^*(X)[Z]\in T_{A^*(X)}\mathcal A$.
The second term therefore vanishes by first-order optimality of
$A^*(X)$. Since $Z$ is arbitrary, the gradient identity follows.

Momentum Transformer introduces momentum into linear attention
\citep{Nguyen2022}, while SympFormer develops inertial Hamiltonian dynamics
on spaces of token distributions \citep{Stein2026}. In our formulation,
the Fisher--Rao geometry acts on the rows of $A$, and attention has
its own evolution equation. A common energy couples this relaxation
to the canonical dynamics of $(X,P)$, yielding the dissipation law
\eqref{eq:coupled-energy}.

\subsection{Two alternative Hamiltonian
formulations}\label{sec:hamiltonian-values}
We present two alternative Hamiltonians incorporating the attention
value map: one through an interaction penalty, the other by setting the
representation velocity equal to the attention output.

\subsubsection{A value-matching energy}
The force in \eqref{eq:score-force} depends only on the query and key
maps. To let the supplied values also guide the representation dynamics
while retaining inertia, we add an attention-weighted squared discrepancy
between $x_i$ and $W_Vx_j$ to the interaction energy. This favours agreement
between each receiving representation and the values to which it assigns
weight. For $\lambda>0$, define $H_\lambda:\mathcal Z\to\R$ by
\begin{equation}\label{eq:value-energy}
\begin{aligned}
H_\lambda(X,P,A)&=\frac{1}{2m}\lVert P\rVert_{\mathrm F}^2+\cF_\lambda(A;X),\\
\cF_\lambda(A;X)&=\cF(A;X)+\frac\lambda2
       \sum_{i,j=1}^N A_{ij}\lVert x_i-W_Vx_j\rVert^2.
\end{aligned}
\end{equation}
The parameter $\lambda$ controls the strength of this value-matching
penalty. With the same canonical and Fisher--Rao structures, the
equations are
\[
\dot X=P/m,\qquad
\dot P=-\nabla_X\cF_\lambda(A;X),\qquad
\dot A=-\gamma\grad_A^{\FR}\cF_\lambda(A;X).
\]
Extending $\cF_\lambda$ in \eqref{eq:value-energy} to positive
matrices and differentiating with respect to $A_{ij}$ gives
\[
\frac{\partial\cF_\lambda}{\partial A_{ij}}
=-\widetilde s_{ij}(X)+\tau(1+\log A_{ij}),\qquad i,j=1,\ldots,N,
\]
where
\[
\widetilde s_{ij}(X)
=s_{ij}(X)-\frac\lambda2\lVert x_i-W_Vx_j\rVert^2.
\]
Thus, at fixed $X$, the equilibrium attention is
$\softmax_{\mathrm{row}}(\widetilde S(X)/\tau)$, where
$\widetilde S=(\widetilde s_{ij})$, and the logit equation is
$\dot U=\gamma(\widetilde S(X)-\tau U)$.

Differentiating the penalty and using row normalisation gives,
for $i=1,\ldots,N$,
\[
-\nabla_{x_i}\cF_\lambda=-\nabla_{x_i}\cF
 +\lambda\left(\sum_{j=1}^N A_{ij}W_Vx_j-x_i\right)
 +\lambda W_V^\top\sum_{k=1}^N A_{ki}(x_k-W_Vx_i).
\]
The first additional term pulls $x_i$ towards its attention-weighted
value average. The second accounts for how changing $x_i$ changes
the values supplied to other tokens. Applying
\eqref{eq:hamiltonian-observable} with $H_\lambda$ from
\eqref{eq:value-energy} gives the energy law
\[
\frac{dH_\lambda}{dt}
=-\gamma\lVert \grad_A^{\FR}\cF_\lambda\rVert_{\FR,A}^2.
\]

\subsubsection{A cotangent lift of the attention velocity}
A different objective is to reproduce the continuous feature velocity
associated with residual updates
$X^{n+1}=X^n+hW_VX^n(A^n)^\top$. To impose the corresponding law
$\dot X=W_VXA^\top$, we add the attention free energy to
the cotangent-lift Hamiltonian and define
$H_{\mathrm{lift}}:\mathcal Z\to\R$ by
\begin{equation}\label{eq:lift-hamiltonian}
H_{\mathrm{lift}}(X,P,A)
=\langle P,W_VXA^\top\rangle_{\mathrm F}+\cF(A;X).
\end{equation}
Its canonical equations, coupled to Fisher--Rao descent in $A$, are
\[
\dot X=W_VXA^\top,\qquad
\dot P=-W_V^\top PA-\nabla_X\cF,\qquad
\dot A=-\gamma\grad_A^{\FR}H_{\mathrm{lift}}.
\]
Extending $H_{\mathrm{lift}}$ in \eqref{eq:lift-hamiltonian}
to positive matrices and differentiating with respect to $A_{ij}$ gives
\[
\frac{\partial H_{\mathrm{lift}}}{\partial A_{ij}}
=p_i^\top W_Vx_j-s_{ij}(X)+\tau(1+\log A_{ij}),
\qquad i,j=1,\ldots,N,
\]
so the attention equation uses scores $s_{ij}(X)-p_i^\top W_Vx_j$.
Here $H_{\mathrm{lift}}$ is linear in $P$: the momentum is an
auxiliary conjugate variable, with no relation $P=m\dot X$.
Applying \eqref{eq:hamiltonian-observable} with $H_{\mathrm{lift}}$
from \eqref{eq:lift-hamiltonian} gives
\[
\frac{dH_{\mathrm{lift}}}{dt}
=-\gamma\lVert\grad_A^{\FR}H_{\mathrm{lift}}\rVert_{\FR,A}^2\leq0.
\]

\subsection{Global well-posedness}\label{sec:coupled-wellposedness}
We now establish global existence and uniqueness for the three coupled
models.

\begin{theorem}[Global well-posedness of the coupled models]
\label{thm:coupled-wellposedness}
Let $\mathcal H$ be one of the Hamiltonians $H$, $H_\lambda$, or
$H_{\mathrm{lift}}$, with $m>0$ for $H$ and $H_\lambda$, and
$\lambda>0$ for $H_\lambda$. For every initial state
$(X^0,P^0,A^0)\in\mathcal Z$, the system
\[
\dot X=\nabla_P\mathcal H,\qquad
\dot P=-\nabla_X\mathcal H,\qquad
\dot A=-\gamma\grad_A^{\FR}\mathcal H
\]
admits a unique smooth solution
\[
(X,P,A):[0,\infty)\longrightarrow\mathcal Z,\qquad
(X(0),P(0),A(0))=(X^0,P^0,A^0).
\]
For each $t\geq0$, the solution depends smoothly on its initial state.
Each attention row remains normalised, and for every finite $T>0$,
\[
\min_{\substack{0\leq t\leq T\\1\leq i,j\leq N}} A_{ij}(t)>0.
\]
\end{theorem}
\begin{proof}
For each Hamiltonian, define its effective score map
$\Sigma:\mathcal X\times\R^{d\times N}\to\R^{N\times N}$ by
\[
\Sigma(X,P)=
\begin{cases}
S(X),&\mathcal H=H,\\
\widetilde S(X),&\mathcal H=H_\lambda,\\
S(X)-P^\top W_VX,&\mathcal H=H_{\mathrm{lift}}.
\end{cases}
\]
The attention derivatives computed above have the common form
$\partial\mathcal H/\partial A_{ij}
=-\Sigma_{ij}(X,P)+\tau(1+\log A_{ij})$.
Applying Lemma~\ref{lem:prelim-logit-flow} to each row therefore gives
\[
A=\softmax_{\mathrm{row}}(U),\qquad
\dot U=\gamma\bigl(\Sigma(X,P)-\tau U\bigr),\qquad
U(0)=\log A^0.
\]
Together with the canonical equations, this defines a smooth vector field
on $\mathcal X\times\R^{d\times N}\times\R^{N\times N}$.
Local ODE theory gives a unique maximal solution $(X,P,U)$ on
$[0,T_{\max})$, with initial state $(X^0,P^0,\log A^0)$.

For each model, the feature and momentum equations are linear in
$(X,P)$ when $A$ is fixed. Their coefficients extend continuously to
the compact set of nonnegative row-normalised matrices, so they are
uniformly bounded. Set
$Q(t)=\lVert X(t)\rVert_{\mathrm F}^2+\lVert P(t)\rVert_{\mathrm F}^2$.
There is a constant $C>0$, depending only on the fixed parameters,
such that
\[
\dot Q(t)
=2\langle X,\dot X\rangle_{\mathrm F}
 +2\langle P,\dot P\rangle_{\mathrm F}
\leq C Q(t).
\]
Gr\"onwall's inequality gives $Q(t)\leq Q(0)e^{Ct}$ for $t<T_{\max}$.
Every entry of $\Sigma$ is quadratic in $(X,P)$, so for some
constant $K>0$,
$\lVert\Sigma(X(t),P(t))\rVert_{\mathrm F}\leq KQ(t)$.
Integrating the logit equation, as in \eqref{eq:attention-solution}, gives
\begin{align*}
U(t)&=e^{-\gamma\tau t}\log A^0
 +\gamma\int_0^t e^{-\gamma\tau(t-s)}\Sigma(X(s),P(s))\,ds,\\
\lVert U(t)\rVert_{\mathrm F}
&\leq\lVert\log A^0\rVert_{\mathrm F}
 +\frac K\tau Q(0)e^{Ct}.
\end{align*}
If $T_{\max}<\infty$, these estimates bound $(X,P,U)$ uniformly
up to $T_{\max}$. Since the lifted vector field is smooth on the whole
Euclidean state space, the ODE continuation theorem would extend the
solution beyond $T_{\max}$. Hence $T_{\max}=\infty$.

The row softmax of $U$ solves the original attention equation and
preserves row normalisation. For
$R_T=\sup_{0\leq t\leq T}\lVert U(t)\rVert_{\mathrm F}<\infty$,
\[
A_{ij}(t)=\frac{e^{U_{ij}(t)}}{\sum_{k=1}^N e^{U_{ik}(t)}}
\geq\frac{e^{-2R_T}}{N}>0,
\qquad 0\leq t\leq T,\quad i,j=1,\ldots,N.
\]
Finally, the original vector field is smooth on $\mathcal Z$, so
uniqueness and smooth dependence on the initial state follow from
smooth ODE theory.
\end{proof}

Theorem~\ref{thm:coupled-wellposedness} concerns the three Hamiltonians
specified above. Additional feature potentials as in
Remark~\ref{rem:feature-potentials} require
appropriate growth or coercivity assumptions for global existence.
Long-time boundedness and convergence require further analysis.

\section{Summary and Outlook}\label{sec:summary}

We developed a variational framework in which softmax minimises
an attention free energy. Its Fisher--Rao gradient flow gives the
entropy-regularised replicator equation \eqref{eq:attention-flow1},
which preserves row normalisation and positivity for positive initial
attention. We proved convergence to the softmax equilibrium
at fixed token representations. For prescribed time-dependent
representations, we introduced a change to logit variables
that linearises the attention dynamics and yields tracking estimates and
sufficient conditions for attention to converge to a fixed matrix.

In Section~\ref{sec:hamiltonian}, we coupled this attention evolution
to Hamiltonian dynamics for the representations and momenta
through a common energy, thereby obtaining
the closed system \eqref{eq:canonical-coupled}.
The canonical Poisson bracket governs the conservative evolution
of $(X,P)$, while the Fisher--Rao gradient flow in $A$ dissipates
the total Hamiltonian. {We also derived a Lagrange--d'Alembert
formulation with Rayleigh dissipation. Finally, we established global
well-posedness for this model and the two alternatives involving the
value map (Theorem~\ref{thm:coupled-wellposedness}).}

{For future work,} numerical simulations could investigate how feature
inertia and finite attention relaxation jointly affect the representations.
{A useful comparison is with the reduced Hamiltonian model that
imposes $A(t)=A^*(X(t))$ at every time, starting from the same initial
representations and momenta. This would help identify how attention
memory and lag affect feature evolution.}

{Stochastic perturbations of the logits provide a way to model
fluctuations while maintaining positive, normalised attention. The
connection between replicator--mutator dynamics and nonlinear filtering
\citep{PathirajaWacker2025} suggests studying how such perturbations affect
tracking
and the coupled energy balance.}

A further direction might involve a {spatial continuum} theory of nonequilibrium
attention. One possibility is to treat attention as an internal
variable, analogous to an order parameter in complex-fluid models.
Such a theory would require specifying the continuum state
variables and a suitable Lie--Poisson structure for their
conservative evolution, together with a continuum Fisher--Rao
gradient flow for attention relaxation.

\section*{Acknowledgements}
We are grateful for thoughtful discussions during this work with C. J.
Cotter, R. Hu, W. Pan, O. D. Street, and C. Tronci. DH is grateful for
partial support from the European Research Council (ERC) Synergy grant
Stochastic Transport in Upper Ocean Dynamics (STUOD) -- DLV-856408.
ABDL is grateful for support from the Universidad de La Laguna through its
Programa Propio de Apoyo a la Investigaci\'on 2026, under the scheme
``Proyectos liderados por personal investigador novel''.

\section*{Use of artificial intelligence}
The authors used large language models to assist with language editing,
corrections, optical character recognition (OCR) of
handwritten iPad notes and other text notes, \LaTeX{} coding, and figure
preparation. The scientific ideas
and results were developed by the authors, who take full responsibility
for the content of the manuscript.

\appendix
\renewcommand{\theequation}{\thesection.\arabic{equation}}
\section{Supporting geometric arguments}\label{sec:geometry-details}
The definitions needed for the dynamics are given in
Section~\ref{sec:preliminaries}. This appendix supplies their supporting
arguments and the entropy interpretations of the Fisher--Rao metric.

\subsection{Induced gradients}\label{app:projection}
\begin{proof}[\normalfont\bfseries Proof of Proposition~\ref{prop:prelim-submanifold-gradient}]
Fix $p\in M$ and $v\in T_pM$. By orthogonality and
Definition~\ref{def:prelim-gradient},
\[
g_p\bigl(\Pi_p\grad_h\overline F(p),v\bigr)
=h_p\bigl(\grad_h\overline F(p),v\bigr)
=d\overline F_p(v)=dF_p(v).
\]
The defining metric duality identifies this vector with $\grad_gF(p)$.
It also proves independence of the ambient extension.
\end{proof}

\subsection{The tangent space of the simplex}\label{app:simplex}
We prove Lemma~\ref{lem:prelim-tangent}.
\begin{proof}
Let $h:U_n\to\R$ be $h(x)=\sum_{j=1}^n x_j$. Its differential
$dh_x(v)=\sum_{j=1}^n v_j$ is surjective at every $x$. Hence
$\mathcal P_n=h^{-1}(1)$ is a regular level set of dimension $n-1$,
with tangent space $T_p\mathcal P_n=\ker dh_p$ at each $p\in\mathcal P_n$.
Differentiating $\sum_{j=1}^n p_j(t)=1$ along a curve in $\mathcal P_n$
gives $\sum_{j=1}^n\dot p_j(t)=0$. Conversely,
for any $v\in\R^n$ with zero sum, $p+tv$ lies in $\mathcal P_n$ for all
sufficiently small $|t|$.
\end{proof}
\subsection{Relative entropy and the Fisher metric}\label{app:kl}
\begin{proposition}[Nonnegativity and local quadratic form]
\label{prop:prelim-kl}
For $p,q\in\mathcal P_n$, one has $\KL(p\|q)\geq0$, with equality
if and only if $p=q$. For fixed $p\in\mathcal P_n$ and $v\in T_p\mathcal P_n$,
\begin{equation*}
\begin{aligned}
\KL(p+\varepsilon v\|p)
&=\frac{\varepsilon^2}{2}g_p(v,v)+O(\varepsilon^3),\\
\KL(p\|p+\varepsilon v)
&=\frac{\varepsilon^2}{2}g_p(v,v)+O(\varepsilon^3),
\end{aligned}
\qquad(\varepsilon\to0).
\end{equation*}
\end{proposition}
\begin{proof}
The inequality $\log x\leq x-1$ gives
\[
-\KL(p\|q)=\sum_{j=1}^n p_j\log\frac{q_j}{p_j}
\leq\sum_{j=1}^n p_j\left(\frac{q_j}{p_j}-1\right)=0.
\]
Equality requires $q_j/p_j=1$ for every $j=1,\ldots,n$.
For sufficiently small $|\varepsilon|$, the vector $p+\varepsilon v$
is positive and normalised. Expanding in the first argument gives
\begin{align*}
\KL(p+\varepsilon v\|p)
&=\sum_{j=1}^n(p_j+\varepsilon v_j)
  \log\left(1+\varepsilon\frac{v_j}{p_j}\right)\\
&=\varepsilon\sum_{j=1}^n v_j
  +\frac{\varepsilon^2}{2}\sum_{j=1}^n\frac{v_j^2}{p_j}
  +O(\varepsilon^3).
\end{align*}
For the second argument,
\begin{align*}
\KL(p\|p+\varepsilon v)
&=-\sum_{j=1}^n p_j
  \log\left(1+\varepsilon\frac{v_j}{p_j}\right)\\
&=-\varepsilon\sum_{j=1}^n v_j
  +\frac{\varepsilon^2}{2}\sum_{j=1}^n\frac{v_j^2}{p_j}
  +O(\varepsilon^3).
\end{align*}
Both linear terms vanish by \eqref{eq:prelim-tangent}, leaving the
Fisher--Rao quadratic form \eqref{eq:prelim-fisher}.
\end{proof}
\subsection{Entropy, the Fisher metric and log-odds}
\label{app:entropy-coordinates}
\begin{proposition}[Entropy, the Fisher metric and log-odds]
\label{prop:entropy-coordinates}
Let $\mathcal E:\mathcal P_n\to\R$ be the negative entropy,
$\mathcal E(p)=\sum_{j=1}^n p_j\log p_j$. In affine coordinates
$z=(p_1,\ldots,p_{n-1})$,
with $p_n=1-\sum_{j=1}^{n-1}z_j$, write
$\widehat{\mathcal E}(z)=\mathcal E(p(z))$. Then
\[
\begin{aligned}
\frac{\partial\widehat{\mathcal E}}{\partial z_j}
&=\log\frac{p_j}{p_n}=\eta_j,\qquad j=1,\ldots,n-1,\\
\frac{\partial^2\widehat{\mathcal E}}{\partial z_j\partial z_k}
&=\frac{\delta_{jk}}{p_j}+\frac1{p_n},\qquad j,k=1,\ldots,n-1.
\end{aligned}
\]
The Hessian is the coordinate matrix of the Fisher--Rao metric.
\end{proposition}
\begin{proof}
Differentiating $p_n=1-\sum_{j=1}^{n-1}z_j$ in the entropy expression
gives the stated derivatives. For the standard basis
$e_1,\ldots,e_n$ of $\R^n$, we have $\partial p/\partial z_j=e_j-e_n$.
Formula~\eqref{eq:prelim-fisher} gives
$g_p(e_j-e_n,e_k-e_n)=\delta_{jk}/p_j+1/p_n$, identifying the Hessian
with the coordinate metric matrix.
\end{proof}

\section{{Residual blocks and continuous depth}}\label{sec:depth}
{We record how scaled residual blocks approximate a differential
equation for the token representations.}

\subsection{{Residual blocks}}\label{sec:blocks}
\begingroup
Consider $L\geq1$ residual blocks, indexed by $\ell=0,\ldots,L-1$.
We retain one unmasked attention head and a feed-forward map, omitting
layer normalisation. For block $\ell$, take matrices
$W_{Q,\ell},W_{K,\ell}\in\R^{d_k\times d}$ and
$W_{V,\ell}\in\R^{d\times d}$. Let $A_\ell^*(X)$ be the softmax
matrix in \eqref{eq:standard-attention} computed with these matrices.
The attention output map is
\begin{equation*}
\mathsf{Att}_\ell:\mathcal X\longrightarrow\mathcal X,
\qquad
\mathsf{Att}_\ell(X)=W_{V,\ell}X\bigl(A_\ell^*(X)\bigr)^\top.
\end{equation*}
Let $\mathsf{MLP}_\ell:\mathcal X\to\mathcal X$ apply the same
smooth feed-forward map $\R^d\to\R^d$ to every column. A residual
block adds the attention output to its input, then adds the feed-forward
output to the resulting representation:
\begin{equation}\label{eq:block}
\mathsf B_\ell
 =(\mathrm{Id}+\mathsf{MLP}_\ell)
   \circ(\mathrm{Id}+\mathsf{Att}_\ell)
 :\mathcal X\longrightarrow\mathcal X.
\end{equation}
Starting from an input feature matrix $X^{(0)}=X^0\in\mathcal X$,
the forward computation is
$X^{(\ell+1)}=\mathsf B_\ell(X^{(\ell)})$ for $\ell=0,\ldots,L-1$.
\endgroup

\subsection{{Continuous-depth scaling}}
The residual block \eqref{eq:block} gives a discrete evolution of the
feature matrix. To compare it with an ordinary differential equation,
introduce a depth spacing $h>0$ and scale both residual increments:
\begingroup
\begin{equation*}
\mathsf B^h_\ell
 =(\mathrm{Id}+h\mathsf{MLP}_\ell)
   \circ(\mathrm{Id}+h\mathsf{Att}_\ell).
\end{equation*}
\endgroup
At $h=1$ this is the block in \eqref{eq:block}.
Writing $Y=(y_1,\ldots,y_N)$, its first stage is
$Y=X+h\mathsf{Att}_\ell(X)$, so
\begingroup
\begin{equation*}
\frac{y_i-x_i}{h}
 =\sum_{j=1}^N A_{\ell,ij}^*(X)W_{V,\ell}x_j,
\qquad i=1,\ldots,N.
\end{equation*}
\endgroup
For the complete block, expanding the composition gives the exact
identity
\begingroup
\begin{equation*}
\begin{aligned}
\mathsf B^h_\ell(X)&=X+hR^h_\ell(X),\\
R^h_\ell(X)
 &=\mathsf{Att}_\ell(X)
   +\mathsf{MLP}_\ell
      \bigl(X+h\mathsf{Att}_\ell(X)\bigr).
\end{aligned}
\end{equation*}
\endgroup

The comparison as $h\to0$ also requires a choice of how the layer maps
vary with depth. Fix $T>0$ and maps
$\mathsf U,\mathsf V:[0,T]\times\mathcal X\to\mathcal X$.
For each spacing $h=T/L_h$, with $L_h$ a positive integer, use
$\mathsf U(t_\ell,\cdot)$ and $\mathsf V(t_\ell,\cdot)$ as the
attention and feed-forward maps of layer $\ell$, where
$t_\ell=\ell h$ and $\ell=0,\ldots,L_h-1$.
Assume that $\mathsf U(t,\cdot)$ and $\mathsf V(t,\cdot)$ have the
attention and feed-forward forms specified in Section~\ref{sec:blocks}.
The maps $\mathsf U,\mathsf V$ are fixed as $h$ changes; the layer maps
are obtained by evaluating them at the depth points $t_\ell$.

For this family, define $\mathsf B_h:[0,T]\times\mathcal X\to\mathcal X$ by
\begin{equation*}
\mathsf B_h(t,X)
 =X+h\mathsf U(t,X)
   +h\mathsf V\bigl(t,X+h\mathsf U(t,X)\bigr).
\end{equation*}
Starting from $X^{(0)}=X^0\in\mathcal X$, the discrete evolution is
$X^{(\ell+1)}=\mathsf B_h(t_\ell,X^{(\ell)})$ for
$\ell=0,\ldots,L_h-1$. The corresponding candidate differential
equation, with $R:[0,T]\times\mathcal X\to\mathcal X$, is
\begin{equation}
\dot X(t)=R(t,X(t)),
\qquad R(t,X):=\mathsf U(t,X)+\mathsf V(t,X),
\qquad X(0)=X^0.
\label{eq:depth-ode}
\end{equation}

Assume that $\mathsf U$ and $\mathsf V$ extend to locally Lipschitz maps
on an open neighbourhood of $[0,T]\times\mathcal X$. Fix a compact
$K\subset\mathcal X$. For sufficiently small $h>0$, the shifted points
$X+h\mathsf U(t,X)$ lie in a fixed compact neighbourhood of $K$.
Local boundedness of $\mathsf U$ and the spatial Lipschitz bound for
$\mathsf V$ then give
\begin{align*}
\mathsf B_h(t,X)-X-hR(t,X)
&=h\bigl[\mathsf V(t,X+h\mathsf U(t,X))-\mathsf V(t,X)\bigr]\\
&=O(h^2),\qquad h\downarrow0,
\end{align*}
uniformly for $t\in[0,T]$ and $X\in K$, in the Frobenius norm.
Thus the scaled residual block agrees with a forward Euler step for
\eqref{eq:depth-ode} up to an $O(h^2)$ remainder.

For $t\in[0,T]$, $X\in K$, and sufficiently small $h>0$ with
$t+h\leq T$, let $\widehat X:[t,t+h]\to\mathcal X$ solve
$\dot{\widehat X}(s)=R(s,\widehat X(s))$ with $\widehat X(t)=X$.
Local boundedness of $R$ gives
$\widehat X(t+r)-X=O(r)$ for $0\leq r\leq h$; its local Lipschitz
continuity in $(t,X)$ therefore gives
\[
\widehat X(t+h)-X-hR(t,X)
=\int_0^h\bigl[R(t+r,\widehat X(t+r))-R(t,X)\bigr]\,dr
=O(h^2).
\]
Combining the two estimates yields the one-step error
$\|\mathsf B_h(t,X)-\widehat X(t+h)\|_{\mathrm F}=O(h^2)$,
with the same uniformity.

\begingroup
\renewcommand{\emph}[1]{#1}
\bibliographystyle{plainnat}
\bibliography{references}
\endgroup
\end{document}